\documentclass[12pt]{article}
\usepackage{placeins}
\usepackage{latexsym,epsfig,graphicx,epstopdf,amsmath,amssymb,amscd,multirow,psfrag,paralist,dsfont,url}
\usepackage[ruled,vlined]{algorithm2e}

\usepackage[titletoc]{appendix}
\usepackage[american]{babel}
\usepackage{varwidth}
\usepackage{verbatim}
\usepackage{bbm}
\usepackage{tikz}
\usetikzlibrary{shapes.geometric, arrows}
\usepackage{ctable}

\usepackage{threeparttable}

\usepackage{algorithm}
\usepackage{algpseudocode}
\usepackage{natbib}
\renewcommand{\arraystretch}{.8}

\begin{document}

\title{A Computationally Efficient Learning of Artificial Intelligence System Reliability Considering Error Propagation}
\author{\large{Fenglian Pan$^1$,  Yinwei Zhang$^2$, Yili Hong$^3$, Larry Head$^2$, and Jian Liu$^{2}$\thanks{Corresponding author: jianliu@arizona.edu}} \\
\normalsize{$^1$Department of Industrial and Systems  Engineering, UNC at Charlotte, Charlotte, NC, USA} \\
\normalsize{$^2$Department of Systems and Industrial Engineering, University of Arizona, Tucson, AZ, USA} \\
\normalsize{$^3$ Department of Statistics, Virginia Tech, Blacksburg, VA, USA}}

\date{}

\maketitle

\begin{abstract}

Artificial Intelligence (AI) systems are increasingly prominent in emerging smart cities, yet their reliability remains a critical concern. These systems typically operate through a sequence of interconnected functional stages, where upstream errors may propagate to downstream stages, ultimately affecting overall system reliability. Quantifying such error propagation is essential for accurate modeling of AI system reliability. However, this task is challenging due to: i) data availability: real-world AI system reliability data are often scarce and constrained by privacy concerns;
ii) model validity: recurring error events across sequential stages are interdependent, violating the independence assumptions of statistical inference; and iii) computational complexity: AI systems process large volumes of high-speed data, resulting in frequent and complex recurrent error events that are difficult to track and analyze. To address these challenges, this paper leverages a physics-based autonomous vehicle simulation platform with a justifiable error injector to generate high-quality data for AI system reliability analysis. Building on this data, a new reliability modeling framework is developed to explicitly characterize error propagation across stages. Model parameters are estimated using a computationally efficient, theoretically guaranteed composite likelihood expectation–maximization algorithm. Its application to the reliability modeling for autonomous vehicle perception systems demonstrates its predictive accuracy and computational efficiency.
\end{abstract}


\newpage
\newpage

\section{Introduction}
\label{sec:introduction}

    Artificial intelligence (AI) systems are becoming increasingly prevalent in various industry sectors, including transportation \citep{mnyakin2023applications}, information technology \citep{chan2019survey}, manufacturing \citep{peres2020industrial}, and healthcare \citep{balagurunathan2021requirements}. While AI technologies have unlocked unprecedented potential for innovation and efficiency in real-world applications, their reliability has emerged as a critical barrier to broader adoption. This challenge is particularly pronounced in safety-critical applications, such as autonomous vehicles (AVs), where failures in AI system may lead to severe consequences. For example, in a recent tragic incident, the AI system of an AV misclassified the white side of a trailer as a bright sky, resulting in a fatal crash \citep{crash}. Such incidents undermines public trust in AI technology and highlight the urgent need for scientifically grounded reliability modeling methods \citep{zou2022modeling}. Motivated by this need, this paper aims to develop a statistical reliability modeling framework, with applications to AI systems in AVs, to enable systematic quantification of AI system performance.

\begin{figure}[t]
        \centering    \includegraphics[width=0.9\textwidth]{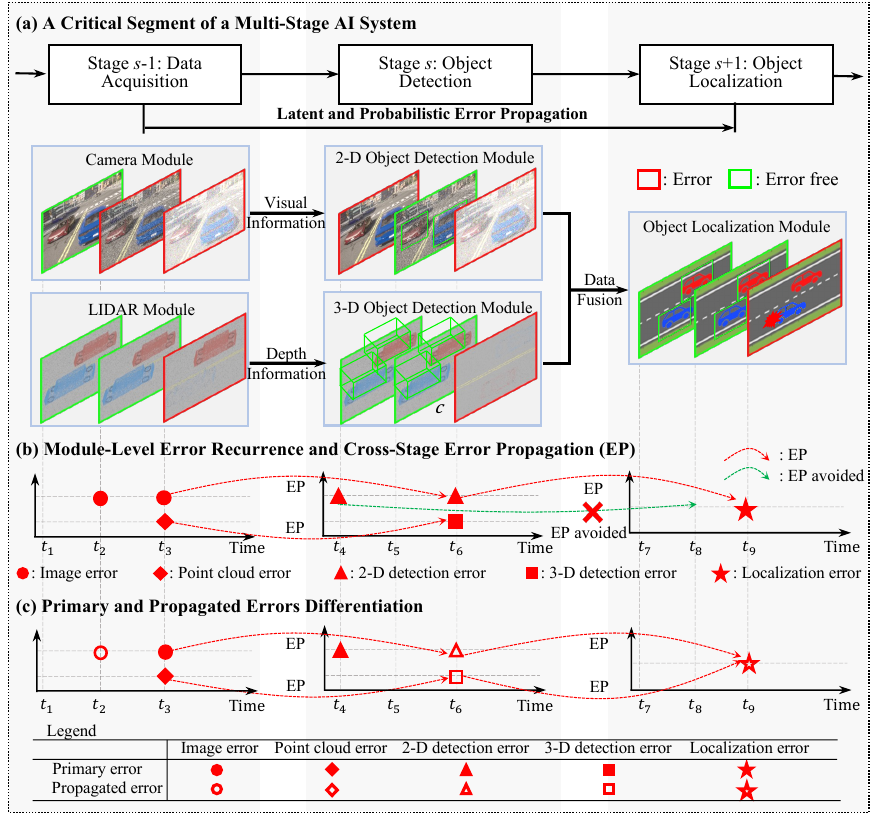}
        \caption{An Illustration of the EP in a Critical Segment of a Multi-Stage AI System}
        \label{fig: Framework}
    \end{figure}

    AI systems in AVs usually integrate multiple sensors and AI/machine learning (ML) algorithms to enable situation awareness, decision-making, and control for AV navigation. In practice, an AI system operates through a series of functionally interconnected stages, where the outputs of one stage serve as the essential inputs for a functionally connected subsequent stage. Within each stage, one or more modules may operate in parallel to provide functional redundancy, ensuring that if one module fails or produces erroneous outputs, the remaining modules can compensate. As shown in Fig. \ref{fig: Framework}(a), the data acquisition stage of a three-stage AI system segment in an AV employs both camera and LiDAR sensors to capture complementary information: high-resolution images and three-dimensional (3-D) point cloud data that ``depict'' surrounding driving environment. These heterogeneous, but complementary, data streams are processed in a subsequent object detection stage, where AI/ML algorithms perform 2-D detection on image data and 3-D detection on LiDAR point clouds to identify vehicles, pedestrians, and other relevant objects, respectively. In the subsequent object localization stage, the detected outputs from both modules are fused to determine precise spatial locations of the objects on the road, enabling the AV to construct an accurate, real-time map of its immediate surroundings. 
    
    While this multi-module, multi-stage structure is critical for ensuring overall system reliability and ensuring the safety of an AV, each module or stage may generate or be exposed to errors under diverse and potentially challenging driving conditions.  For example, in Fig. \ref{fig: Framework}(a), the camera module may capture low-quality images in adverse weather conditions, such as heavy rain or dense fog. The 2-D object detection module may fail to detect the car on the road due to the inherent limitations of its own algorithm, or to the degraded quality of the input image. Such error events tend to recur intermittently over time and can be represented as random points along the time domain, as shown in Fig. \ref{fig: Framework}(b). Individually, these errors may not immediately result in catastrophic failure of the AI system. However, due to the functional dependencies among stages, some errors may propagate to downstream stages and trigger additional errors in subsequent stages. This phenomenon is referred to as error propagation (EP) in the rest of this paper. As illustrated in Fig. \ref{fig: Framework}(b), a low-quality image error at time $t_3$ in the data acquisition stage propagates to 2-D object detection stage and causes 2-D detection error at time $t_6$, which further propagates to localization stage and lead to localization error at time $t_9$. To better understanding EP, this paper distinguishes two types of errors, defined as follows. 
    \begin{itemize}
            \item \textbf{Definition 1.} A \textit{Primary Error} is an error event caused by nonconforming performance of a module at a certain functional stage itself, which is independent of the performance/input of modules in upstream stages. For example, in Fig. \ref{fig: Framework}(c), even with a high-quality image captured by a camera module in data acquisition stage at $t_1$, the 2-D object detection module still failed to detect the object in the scene at $t_4$, indicating that this mis-detection error event arises directly (or ``primarily'') from the inherent limitation of the object detection algorithm. Such primary errors are marked as solid-filled points in Fig. \ref{fig: Framework}(c).
        \item \textbf{Definition 2}. A \textit{Propagated Error} is an error event triggered by erroneous input from an upstream stage. For instance, in Fig. \ref{fig: Framework}(c), the 2-D object detection error at time $t_6$ is triggered by an excessively noisy image captured by the camera sensor module in the data acquisition stage at time $t_3$. Such propagated errors are marked as hollow-filled points in Fig. \ref{fig: Framework}(c).
    \end{itemize}

    It is essential to explicitly model EP for AI system reliability analysis. This is because EP effects may accumulate across stages over time, potentially degrading system performance and ultimately leading to failure of the entire AI system. Neglecting such effects may result in an inaccurate system reliability model that underestimates the risk of system-level failure. Nevertheless, modeling EP in AI systems presents several unique challenges, including:
  \begin{enumerate}[(1)]
      \item \textit{Data challenge: lack of readily available AI reliability data.} A fundamental challenge in statistical AI system reliability modeling and analysis is the scarcity of reliability data. This is especially true for rare system-failure event data, such as AV disengagement event (not necessarily fatalities) \citep{min2022reliability}, which would require real-world tests with vehicles driven for hundreds of millions of miles to collect \citep{kalra2016driving} \citep{zheng2025dr}. In contrast, the AV simulation can offer a cost-effective and highly repeatable means of data generation. Simulation platforms, such as Automated Driving Toolbox, CARLA, and Autoware, can provide a variety of driving environments, traffic agents (e.g., vehicles and pedestrians), and rule-based traffic flows \citep{li2023modified}. Nevertheless, concerns regarding the fidelity, realism, and representativeness of simulated scenarios relative to real-world driving conditions continue to limit the reliability of simulation data. 
      \item \textit{Model challenge: latent and probabilistic EP.} The presence of EP introduces several fundamental challenges for AI system reliability modeling. First, EP makes the error events across stages NOT independent given all of the known factors, which violates the independence assumptions of statistical reliability modeling. Second, EP is inherently latent, i.e., downstream errors do not carry labels indicating whether they are primary or propagated, making these error types indistinguishable given the observed data.  Third, EP itself is probabilistic rather than deterministic. An upstream error may increase the risk of downstream failures, but does not guarantee their occurrence. These characteristics jointly complicate EP modeling in AI systems.
      \item \textit{Inference challenge: inaccurate and inefficient EP estimation.} The latent and probabilistic nature of EP tends to confound primary errors with propagated ones, thereby obscuring the original source of system failures and leading to biased model inference. Furthermore, the scale of recurring error event data collected across all system modules and stages, spanning the entire system operational duration, is vast. Performing inference on such datasets while simultaneously differentiating the effects of EP creates a significant computational bottleneck for traditional estimation methods.
\end{enumerate}

 
 To address the data challenge, existing studies have proposed collecting AI system reliability data from real-world testing or simulation, as summarized in Table \ref{tab:Dataset}. Most publicly available AI system reliability data are collected independently at either the module or system level. For example, at the module level, the KITTI dataset \citep{geiger2013vision} provides reliability data of multiple AI system modules in moving AVs on the road, e.g., camera images, laser scans, GPS measurements, and IMU accelerations. Similarly, the classification errors were collected from an image recognition model to assess the reliability of ML algorithms \citep{Lianetal2021Robustness}, \citep{Faddietal2024}. At the system level, the Autonomous Vehicle Tester (AVT) program \citep{CAdriving} reports disengagement event data collected during real-world testing of AV systems, providing valuable insights into AVs reliability. Additionally, online database \citep{AIIncidentDB} documents incidents involving the use of AI systems that result in harm or near-harm consequences. However, because cross-stage functional connections are absent, such disjoint module-level or system-level data alone cannot be used to model the EP.   
\begin{table}[htbp]
\centering
\caption{Existing AI System Reliability Data}
\label{tab:Dataset}
\begin{tabular}{llll}
\hline 
\hline
\multicolumn{1}{c}{\textbf{Dataset}}& \multicolumn{1}{c}{\textbf{Level}}  & \multicolumn{1}{c}{\textbf{Source}} \\
\hline
KITTI {\citep{geiger2013vision}}  & Module  & Real-world\\
\hline
ML/AI Algorithm Error \citep{Lianetal2021Robustness}  \citep{Faddietal2024} &  Module  & Simulation\\
\hline
AV Disengagement \citep{CAdriving} & System & Real-world \\
\hline
General AI Incidence Data \citep{AIIncidentDB} & System  & Real-world\\
\hline
\end{tabular}
\end{table}

  To address the reliability modeling challenge, various stochastic approaches have been developed to model different types of reliability data, such as time-to-event data and degradation data \citep{gorjian2010review}. For example, a Bayesian hazard model was introduced to consider latent heterogeneity in lifetime data \citep{li2016bayesian}, while a Bayesian nonparametric model was proposed to model the heterogeneous time-to-event data \citep{li2017bayesian}. For degradation data, Wiener-process-based methods \citep{zhang2018degradation}, Gamma process models \citep{pan2011reliability}, and Bayesian models \citep{yuan2019reliability} have been widely employed.  In the context of AI reliability analysis, \citep{wang2020deep} proposed a new data fusion model that utilizes a set of degradation data to monitor the reliability of AI systems. These methods primarily focus on single failure events or continuous degradation until failure and are well-suited for modeling lifetime and degradation data. However, they are not directly applicable to recurrent-event data, which involves modeling events that occur over time. To analyze recurrent event data, approaches such as the Homogeneous Poisson Process (HPP) \citep{hossain1993estimating, yang2024measurement}, the Non-homogeneous Poisson Process (NHPP) \citep{pham2003nhpp}, and the Renewal Process (RP) \citep{pyke1961markov} are commonly employed. Among these methods, the NHPP models have particularly gained much popularity. For instance, the NHPP has been developed to model recurrent disengagement events observed during autonomous vehicle driving tests \citep{min2022reliability}. Similarly, software reliability growth models (SRGMs) \citep{8987509} have been built using the same dataset to analyze the AV reliability over time. Being successful in predicting module-level or system-level reliability, however, these methods didn't consider the event-level interdependency between different module/system  (i.e., the cross-stage EP). To address this limitation, a multi-stage Hawkes Process (MSHP) \citep{pan2022quantifying} was recently proposed to explicitly quantify the EP in a multi-stage serial perception system in AV. This method assumes only one module in each functional stage and cannot model scenarios with multiple modules within a stage. A subsequent research \citep{pan2024reliability} introduced an event-triggering point process that extends the model to handle such scenarios. Despite these advancements, how to efficiently estimate the model parameter from massive reliability data has not been investigated in these models.

To address the inference challenge, the likelihood-based methods have been widely used to estimate the model. A well-established approach is maximum likelihood estimation (MLE) \citep{pan2025modeling}, which is popular for its asymptotic efficiency and desirable statistical properties. However, MLE can pose significant challenges in terms of convergence and computational efficiency, particularly when estimating models with a large number of parameters \citep{kapur2014reliability}. These difficulties are further compounded in the presence of incomplete data, where the need to account for unobserved or latent information (e.g., the latent EP in this paper) adds additional layers of mathematical and computational complexity. To address these limitations, researchers have explored alternative ways to implement MLE in settings with incomplete or partially observed data. One of the most widely used approaches in such contexts is the expectation–maximization (EM) algorithm, which is designed to handle latent information by iteratively alternating between two steps: (1) the expectation (E) step, which computes the expected log-likelihood given the observed data and current parameter estimates, and (2) the maximization (M) step, which updates the parameters by maximizing this expected log-likelihood \citep{dempster1977maximum}. The EM algorithm was first applied to estimate the parameters of SRGM in \citep{okamura2003iterative} for the Exponential, Gamma, and Weibull distributions, and then later in \citep{okamura2013application} for other fault detection time distributions. However, the update rule for the EM algorithm for all distribution parameters can be computationally complex when applied to a large dataset \citep{zeephongsekul2016maximum}. In this context, some studies have explored an extensions of the EM algorithm to pseudo-likelihood settings, including pseudo-EM methods for network tomography \citep{liang2003maximum}, pairwise EM approaches for spatial generalized linear mixed models \citep{varin2005pairwise}, and composite likelihood EM formulations for multivariate hidden Markov models \citep{gao2011composite}. These developments primarily focus on latent state sequence models or high-dimensional correlated data. To date, there has been limited work on developing composite likelihood EM methods for recurrent-event data, particularly for models that incorporate latent event-propagation mechanisms.

This paper develops a comprehensive methodology to address challenges in data, modeling, and estimation for AI system reliability, with explicit consideration of EP. First, we develop an error injection (EI) mechanism within a scalable physics-based simulation platform to systematically generate and collect recurring error event data across multiple interconnected stages of AI systems. Building upon these data, we propose an intensity-decomposition approach that explicitly models latent and probabilistic EP by mathematically distinguishing the effects of primary and propagated errors. To efficiently facilitate inference under latent and probabilistic EP, we introduce latent variables that indicate error types and design a composite likelihood–based EM algorithm. The proposed approach can achieve higher computational efficiency compared to conventional full-likelihood EM methods while preserving theoretical properties. The main contributions of this research include:

\begin{itemize}
\item \textbf{A systematic data collection framework,} which systematically generates and collects reliability data across multiple interconnected stages of an AI system. The framework enables targeted perturbations in specific modules at user-defined times and probabilities. By incorporating real-world testing information, it helps bridge the gap between simulation and real-world testing, providing more realistic and justifiable reliability data.
\item \textbf{An intensity-decomposition-based EP modeling approach,} which explicitly captures latent and probabilistic EP by explicitly differentiating primary and propagated errors and decomposing the overall error effects into interpretable intensity components, thereby relaxing unrealistic independence assumptions across system modules.
\item \textbf{A new CLEM algorithm for point process model inference}, which exploits localized composite likelihood construction and substantially reduces computational cost compared with the conventional full-likelihood EM approach. A stepwise Friedman Test selection procedure is further introduced to guide the construction of the composite likelihood, ensuring that estimation accuracy remains comparable to that of the standard EM algorithm. Moreover, the proposed framework preserves key theoretical properties, including the ascent property and statistical consistency.
\end{itemize}

Table 2 summarizes the limitations of the existing methods and the features of the proposed methods in addressing the data, model, and inference challenges in AI system reliability modeling

\begin{table*}[ht]
\centering
\small
\caption{Challenges and Solutions: Existing vs. Proposed Methods}
\renewcommand{\arraystretch}{1.25}
\begin{tabular}{>{\centering\arraybackslash}m{2.4cm}|>{\arraybackslash}m{6.8cm}|>{\arraybackslash}m{6.3cm}}
\hline\hline
\textbf{Challenges} & \textbf{Limitations of Existing Approaches} & \textbf{Features of Proposed Methods} \\ \hline

\multirow{3}{*}{\textbf{Data}} 
& Disjoint module- or system-level data without EP information 
& Systematic data collection with EP information \\ \cline{2-3}

& Expensive and time-consuming real-world tests with limited scenarios
& Scalable, EI-enabled simulation with diverse driving scenarios \\ \cline{2-3}
\hline

\textbf{Model} 
& Potential EP effects neglected
& EP effects explicitly modeled as a quantifiable hazard intensity component \\ \hline

\textbf{Inference} 
& Computationally prohibitive/inefficient
& Computationally efficient with theoretical guarantee \\ \hline\hline

\end{tabular}
\label{tab:challenges_methods}
\end{table*}

The remainder of this paper is organized as follows. Section \ref{sec:Methodology} introduces a methodology. Section \ref{sec: Case Study} demonstrates the advantages of the proposed method based on the numerical case study and physics-based simulation case study. The conclusion and future work will be given in Section \ref{sec:Conclusions}. 

\section{Methodology}
\label{sec:Methodology}

  The objective of this study is to effectively and efficiently analyze the reliability of AI systems in AVs, with a particular focus on accurate EP effect quantification. This analysis relies on extensive reliability data and subsequent modeling and estimation techniques, which can be implemented in an integrative methodological framework presented in this section, including data simulation, model formulation, parameter estimation, and performance evaluation.
  
\subsection{Data Simulation with a Justifiable EI-Enabled Framework}
\label{subsec:Data Generation}

\begin{figure}[t]
\centering
\includegraphics[width=0.9\textwidth]{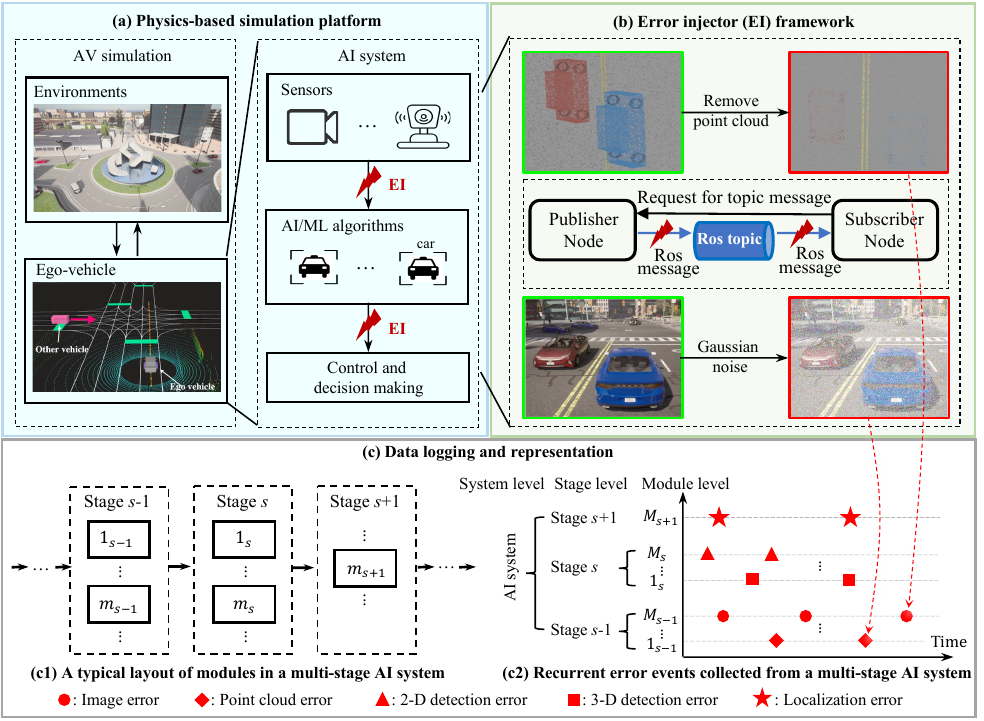}
    \caption{An Illustration of Error Injection Framework in a Physics-Based Simulation Platform}
    \label{fig: Error Injection Framework}
\end{figure}

To address the data challenge by collecting reliability data with EP information, we propose a physics-based simulation framework to systematically generate AI system reliability data for AVs under diverse traffic scenarios. As illustrated in Fig. \ref{fig: Error Injection Framework}, the framework comprises three main components.
\begin{itemize}
    \item \textbf{A physics-based AV simulation platform}, which is used to simulate AV performance in diverse traffic scenarios. As illustrated in Fig. \ref{fig: Error Injection Framework} (a), this platform consists of two main elements: (i) \textit{environment}, which integrates diverse physical models, including infrastructures, driving scenarios, and traffic-related agents, facilitating a high-fidelity simulation environment that closely mirrors real-world driving conditions; (ii) \textit{ego vehicle}, which interacts with the driving environment via an AI system. This AI system fuses data from multiple sensors (e.g., camera, RADAR, and LiDAR) to perceive the surroundings, leveraging AI/ML algorithms for object detection, localization, and path planning.
    \item \textbf{An EI-enabled framework}, which is built on the Robot Operating System (ROS) \citep{quigley2009ros} that adopts a publisher-subscriber mechanism to facilitate data transmission between modules at upstream (publisher) and downstream (subscriber) stages in an AI system. With this mechanism, sensors and perception modules act as upstream publishers, continuously generating data streams such as camera images, LiDAR point clouds, or radar signals. These data streams are published as topics in ROS. Downstream modules, such as object detection, tracking, and sensor fusion nodes, function as subscribers. They receive relevant topics, process the incoming data, and produce higher-level perception outputs such as detected objects, lane boundaries, or free-space maps.  As illustrated in Fig. \ref{fig: Error Injection Framework} (b), the proposed EI framework involves the following three key steps: (i) create a Publisher Node to transmit data from the upstream modules to downstream modules through ROS topics, (ii) inject realistic errors (e.g., randomly remove the point cloud data or add Gaussian noise to the image data) into the ROS topics, and (iii) create a subscriber node to receive erroneous data in downstream modules. This design allows for targeted EI into specific AI modules at user-defined timestamps and probabilities. Formally, given a module $m_s$, a user-defined timestamp $t^{err}_{m_s}$, and a probability $p_{t^{err}_{m_s}}$, the EI indicator function is defined as $f(m_s, t^{err}_{m_s}, p_{t^{err}_{m_s}}) = \mathbbm{1}\{u \leq p_{t^{err}_{m_s}}\}, u \sim U(0, 1)$, where $u$ is a random number drawn from a uniform distribution, and $\mathbbm{1}\{\cdot\}$ is an indicator function. Here, $f(m_s, t^{err}_{m_s}, p_{t^{err}_{m_s}}) = 1$ indicates that an error is successfully injected into module $m_s$ at time $t^{err}_{m_s}$ with probability $p_{t^{err}_{m_s}}$, whereas $f(m_s, t^{err}_{m_s}, p_{t^{err}_{m_s}}) = 0$ indicates that an error is not successfully injected into module $m_s$ at time $t^{err}_{m_s}$. Fig. \ref{fig: Error Injection Framework}(b) shows a successful EI application in the perception module, where clear camera images and dense LiDAR point clouds (blue and red car) are transformed into noisy images and sparse point clouds. With this EI framework, we can design the EI probabilities based on empirical findings from real-world tests, which can narrow the gap between the simulation and reality. In addition, for rare events, instead of waiting for errors to occur naturally, the EI can accelerate the error occurrence process, providing an efficient method to generate and log sufficient error event data for reliability modeling.  
    \item \textbf{A data logging component}, which is responsible for systematically collecting recurrent error events from multiple modules across different functional stages throughout the continuous operation of the AV. Fig. \ref{fig: Error Injection Framework}(c1) shows a representative layout of a three-stage segment of a multi-stage AI system, where $s$ is the stage index and $m_s$ is the module index at stage $s$. The recorded error events can be organized hierarchically at module, stage, and system levels, as illustrated in Fig. \ref{fig: Error Injection Framework}(c2). For each error event, the logging component captures key information, including its error type, occurrence time, stage index, module index (see Subsection \ref{subsec: Model Formulation} for notation details). By jointly tracking error events across stages over time, the logged data captures the temporal and cross-stage dependencies among modules, thereby providing the necessary information to characterize and model EP throughout the multi-stage AI system.
\end{itemize}

\subsection{Model Formulation}
\label{subsec: Model Formulation}
\begin{figure}
\centering
    \includegraphics[width=0.9\textwidth]{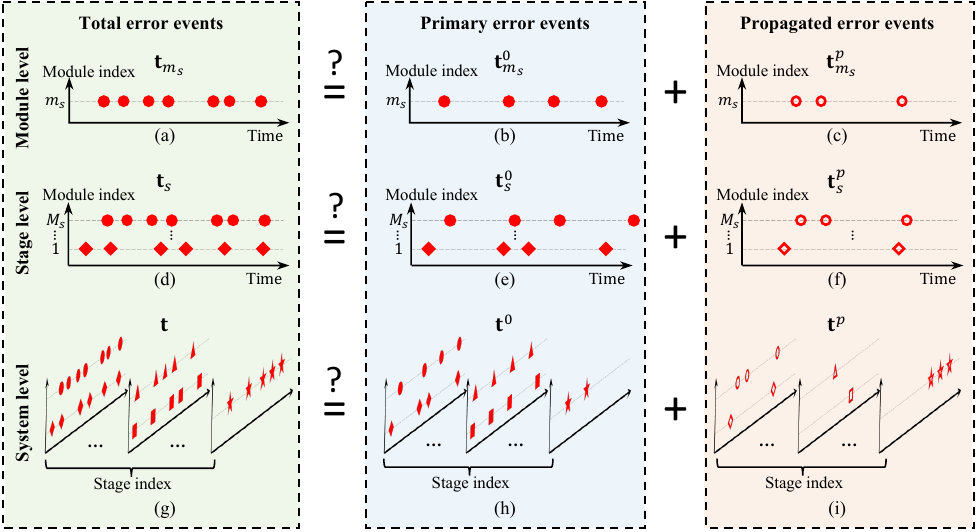}
    \caption{Illustration of Error Event Differentiation}
    \label{fig: EP analysis}
\end{figure}

Consider a multi-stage AI system with $S$ functional stages, each of which fulfills its functionality with $M_s$ module(s), where $M_s\geq 1$. Let $m_s \in \{1_s,...,M_s\}$ denote the index of the $m_s$-th module at stage $s$, where $s\in \{1,2,...,S\}$. At the module level, let $\mathbf{t}_{m_s}= [t_{m_s^1}, ..., t_{m_s^i}, ..., t_{m_s^{n_{m_s}}}]^\top\in \mathbb{R}^{n_{m_s}\times1}$ denote the vector of observed error events times collected from module $m_s$. Here, $t_{m_s^i}$ is the timestamp of the $i^{\text{th}}$-th error event in module $m_s$ and $n_{m_s}$ is the number of error events observed in module ${m_s}$, as illustrated in Fig. \ref{fig: EP analysis} (a). At the stage level, the collection of error events observed in stage $s$ is represented by $\mathbf{t}_s=\cup_{m_s=1}^{M_s}\mathbf{t}_{m_s}$ and the overall system-level error events are represented by $\mathbf{t}= \cup_{s=1}^{S}\mathbf{t}_{s}$, as shown in Fig. \ref{fig: EP analysis} (d) and (g), respectively. The error events, $\mathbf{t}$, collected from an AI system can be naturally characterized as a multivariate point process evolving over time \citep{cox1980point}. In general, a point process is described by its conditional intensity function (CIF, ``\textit{intensity}" hereafter), which quantifies the instantaneous probability of an event occurring at time $t$, given the event history up to that time. Mathematically, the module-level intensity associated with module $m_s$ is defined as, 
 \begin{equation}
    \lambda_{m_s}(t|\mathbf{t}_{m_s})=\lim_{dt \to 0} \frac{E[N(t+dt)-N(t)|\mathbf{t}_{m_s}]}{dt}, 
 \end{equation}
 
 \noindent where $N(t)$ denote the cumulative number of events up to time $t$, and $E[N(t+dt)-N(t)|\mathbf{t}_{m_s}]$ is the expected number of new events in the infinitesimal interval $(t, t+dt]$ conditional on the historical $\mathbf{t}_{m_s}$. This intensity function provides a principal way to model the stochastic occurrence of errors at a given module $m_s$. It serves as the fundamental building block of the proposed EP model. 

It is worth noting that the raw error event observations in Fig. \ref{fig: EP analysis} (a), (d), and (g) do not carry labels that indicate whether they are primary errors or propagated errors. In other words, EP is latent in the observed data. To explicitly model this latent EP, this paper distinguishes the module-level error events $\mathbf{t}_{m_s}$ can be differentiated into two categories: (i) primary errors, denoted by $\mathbf{t}_{m_s}^0$, and propagated errors, denoted by $\mathbf{t}_{m_s}^p$. One illustrative example of this distinction is shown in Fig.  \ref{fig: EP analysis} (a)-(c). Similarly, stage-level error events $\mathbf{t}_s$ can be differentiated into primary errors, $\mathbf{t}_s^0$, and propagated errors, $\mathbf{t}_s^p$, as illustrated in Fig. \ref{fig: EP analysis} (d)–(f). System-level errors, $\mathbf{t}$, can be differentiated into the primary error, $\mathbf{t}^0$, and the propagated error, $\mathbf{t}^p$, as shown in Fig. \ref{fig: EP analysis} (g)–(i). Based on these event decompositions, this paper decomposes the total error intensity of module $m_s$ as the sum of a primary error intensity and an accumulated propagated error intensity caused by stage $s-1$. To make this decomposition interpretable and mathematically tractable, we made two assumptions regarding the propagation mechanism as below.
\begin{itemize}
    \item \textbf{Assumption 1.} Errors in module $m_s$ can be propagated \textit{only} from its immediate preceding stage, which accumulates the propagation effects from further upstream stages.
    \item \textbf{Assumption 2.} Given the external environmental factors, such as rainy and/or windy weather, the error events collected from the parallel modules within the same stage are independent.
\end{itemize}

Under these two assumptions, the total intensity of module $m_s$ can be decomposed as
\begin{equation}
 \label{eq:Total intensity}
    \lambda_{m_s}(t|\mathbf{t}_{m_s})
    = \lambda_{m_s}^0(t|\mathbf{t}_{m_s}^0) +  \sum\nolimits_{m_{s-1}=1}^{M_{s-1}} \lambda_{m_s, m_{s-1}}^p(t|\mathbf{t}_{m_{s-1}}^p),
\end{equation}

\noindent where the first term, $\lambda_{m_s}^0(t|\mathbf{t}_{m_s}^0)$, denotes the primary error intensity of module $m_s$ and the second term, $\sum\nolimits_{m_{s-1}=1}^{M_{s-1}} \lambda_{m_s, m_{s-1}}^p(t|\mathbf{t}_{m_{s-1}}^p)$, represents a collection of propagated error intensities inherited from the immediate proceeding stage $s-1$. Specifically, the primary error intensity, $\lambda_{m_s}^0(t|\mathbf{t}_{m_s}^0)$, characterizes the instantaneous occurrence rate of errors that causes by module $m_s$ itself, independent of upstream modules. It reflects the intrinsic reliability of the module 
$m_s$ without considering the EP. The propagated error intensity, $\lambda_{m_s, m_{s-1}}^p(t|\mathbf{t}_{m_s}^p)$, captures the instantaneous occurrence rate of errors at module $m_s$ that are probabilistically triggered by errors occurring in an upstream module $m_{s-1}$. This term explicitly explains the functional dependencies between module $m_{s-1}$ and module $m_s$ and provide a mechanism for modeling EP throughout the multi-stahe AI system.

Appropriate parameterization of both the primary error intensity and the propagated error intensity in Eq. (\ref{eq:Total intensity}) is critical for accurately modeling the AI system reliability. This paper assumes that, without considering EP, the primary error intensity of module $m_s$ is a constant over time, i.e., 
     \begin{equation}
 \label{eq:Baseline intensity}
    \lambda_{m_s}^0(t | \mathbf{t}_{m_s}^0) = \lambda_{m_s}^0,  \text{  }  \lambda_{m_s}^0 > 0.
\end{equation} 
This is because AVs are subject to strict safety and reliability standards, such as ISO 26262 \citep{kafka2012automotive}, which emphasize minimizing variability in error rates to ensure safe operation. To meet these standards, AI modules in AVs undergo rigorous development, testing, and validation processes, all aimed at delivering consistent performance under normal operating conditions (i.e., without considering EP). 

For the propagated error intensity, we adopt a different characterization. We assume that an error that occurs in an upstream module tends to propagate immediately to downstream modules. This effect naturally diminishes over time because the vehicle’s decision-making process is primarily driven by the most recent input data. As new information continuously arrives, earlier errors are corrected, filtered, or overridden, thereby reducing their long-term influence on the system. To capture this mechanism, we model the propagated error process as a function of the upstream error process, modulated by an exponential decay function, i.e.,

\begin{equation}
\label{eq: propagated intensity}
    \lambda_{m_s, m_{s-1}}^p(t|\mathbf{t}_{m_{s-1}}^p)  = \sum_{t_{m_{s-1}^j}<t}  \alpha_{m_s, m_{s-1}} \cdot \exp[-\beta_{m_s, m_{s-1}} \cdot (t-t_{m_{s-1}^j})],
\end{equation}
    \noindent where parameter $\alpha_{m_s,m_{s-1}}$ models the probabilistic nature of latent EP, with $\alpha_{m_s,m_{s-1}} = 0$ indicates that no EP exists between from module $m_{s-1}$ and module $m_s$, whereas a positive $\alpha_{m_s,m_{s-1}}$ quantifies instantaneous EP effect for errors occurred in module $m_{s-1}$ to propagate to module $m_s$. Parameter $\beta_{m_s,m_{s-1}}$ models the temporal decay rate of the EP effects from module $m_{s-1}$ to module $m_s$. Both $\alpha_{m_s,m_{s-1}}$ and $\beta_{m_s,m_{s-1}}$ are uniquely associated to a module pair, ($m_{s-1}$, $m_s$), characterizing the latent and probabilistic EP between module $m_{s-1}$ and $m_s$. A higher $\alpha_{m_s,m_{s-1}}$ indicates a greater instantaneous EP effect, and a lower $\beta_{m_s,m_{s-1}}$ indicates a lasting EP effect.

\subsection{Model Estimation}
\label{sec: Model Estimation}

 With the model established, this section focuses on developing a likelihood-based estimation algorithm to estimate parameters defined in Eqs. (\ref{eq:Total intensity})-(\ref{eq: propagated intensity}). This algorithm is designed to: (i) explicitly distinguish and quantify primary errors and propagated errors, (ii) improve computational efficiency, and (iii) demonstrate theoretical guarantees for convergence and consistency.
 
\subsubsection{Likelihood Function}
\label{subsec: Likelihood}

Let $\boldsymbol{\Theta}_{m_s} = \big\{\lambda^0_{m_s}, \, \alpha_{m_s,m_{s-1}}, \, \beta_{m_s,m_{s-1}} : m_{s-1}=1,\dots,M_{s-1}\big\}$ denote the parameter set of module $m_s$. The complete parameter set of an AI system can be defined as $\boldsymbol{\Theta} = \{\boldsymbol{\Theta}_{m_s}: s=1, \cdots, S, m_s =1,\cdots,M_s\}$.

\noindent \textbf{Proposition 1}. Let $T$ denote the length of the observation window for all modules, the overall log-likelihood across all the modules and stages in an AI system will be:

   \begin{equation}
 \label{eq:Overall log-likelihood}
\ell(\mathbf{\Theta}) = \sum_{s=1}^{S}\sum_{m_s=1}^{M_s}\ell_{m_s}(\mathbf{\Theta}_{m_s}\mid \mathbf{t}_{m_s}),
    \end{equation}

\noindent where $\ell_{m_s}$ is the log-likelihood for module $m_s$ and can be formulated as
 
\begin{equation}
 \label{eq:log-likelihood}
\ell_{m_s}(\mathbf{\Theta}_{m_s}|\mathbf{t}_{m_s}) = \sum_{i=1}^{n_{m_s}}\log(\lambda_{m_s}(t_{m_s^i})) - \int_0^T \lambda_{m_s}(t)dt.
\end{equation}

\noindent $\lambda_{m_s}(t)$ is defined in Eqs. (\ref{eq:Total intensity}) - (\ref{eq: propagated intensity}). The parameters $\boldsymbol{\Theta}$ can be estimated by finding the values of $\hat{\boldsymbol{\Theta}}$ that maximizes $\ell(\mathbf{\Theta})$ in Eqs. (\ref{eq:Overall log-likelihood}) - (\ref{eq:log-likelihood}). Standard numerical routines for maximum likelihood estimation (MLE) can be directly applied to obtain parameter estimates. However, it can be observed that we need to simultaneously estimate $M_1 + \sum_{s=2}^S{M_s}(2\times M_{s-1}+1)$ parameters in this model, where $M_s, s=1, ..., S$, denotes the number of modules at stage $s$. Even for a moderate $M_s$, such high-dimensional optimization can become computationally intensive and numerically unstable \citep{veen2008estimation}. Moreover, beyond computational challenges, directly applying MLE does not distinguish between the primary errors $\mathbf{t}^0$ and propagated errors $\mathbf{t}^p$, which limits its ability to quantify the latent and probabilistic EP within the system.

\subsubsection{Latency Indicator}
\label{sec:Latency Indicator}

To distinguish between primary and propagated errors, we introduce a latency indicator variable, $I_{m_s^i}$, for each error event $t_{m_s^i} \in \mathbf{t}_{m_s}$. $I_{m_s^i}$ is defined as
\begin{equation}
\label{eq:Latency Indicator}
  I_{m_s^i} =
    \begin{cases}
      0, & \text{if $t_{m_s^i}$ is a primary error}, \\[6pt]
      j, & \text{if $t_{m_s^i}$ is triggered by an upstream error $t_{m_{s-1}^j} \in \mathbf{t}_{s-1}$},
    \end{cases}    
\end{equation}

\noindent where, $j \neq 0$ denotes the index of an error event in module $m_{s-1}$ in the previous stage $s-1$. Let $\mathbf{I}$ denote the collection of all latency indicators, i.e., $\mathbf{I} = \big\{ I_{m_s^i} : s=1,\dots,S;\; m_s=1,\dots,M_s;\; i=1,\dots,n_{m_s} \big\}$. Given $\mathbf{I}$, the error events $\mathbf{t}$ can be decomposed into two disjoint subsets,  i.e., $ \mathbf{t} = \mathbf{t}^0 \cup \mathbf{t}^p$, where $\mathbf{t}^0 = \{t_{m_s^i}\in\mathbf{t}|I_{m_s^i} = 0\}$ and $\mathbf{t}^p = \{t_{m_s^i}\in\mathbf{t}|I_{m_s^i} = j\}$.

Inference of the latency indicators $\mathbf{I}$ is critical for the model estimation based on the likelihood functions defined in Eqs. (\ref{eq:Overall log-likelihood})-(\ref{eq:log-likelihood}). By treating $\mathbf{I}$ as missing values, the estimation can be implemented with an EM algorithm, as described in \citep{pan2024reliability}. However, the EM algorithm will be computationally prohibitive when the number of error events is large. This is because the decay kernel function in Eq. (\ref{eq: propagated intensity}), $\exp[-\beta_{m_s, m_{s-1}} \cdot (t-t_{m_{s-1}^j})]$, with infinite time supports imply that every historical error event in upstream stages could potentially propagate to the downstream stages and contribute to their error intensity functions. Thus, in the E-step of the EM algorithm, evaluating the distribution of each latency indicator $I_{m_s^i}$ requires looping over the complete history of upstream error events $t_{m_{s-1}^j} \in \mathbf{t}_{s-1}$. It involves computing a series of conditional probabilities: $p_{m_s^i}^{(0)} = \Pr(I_{m_s^i} = 0)$, representing the probability that $t_{m_s^i}$ is a primary error,  
and $p_{m_s^i}^{(j)} = \Pr(I_{m_s^i} = j)$, representing the error $t_{m_s^i}$ is a propagated error triggered by the upstream error $t_{m_{s-1}^j}$. As the observation window expands, the number of such probabilities grows quadratically with the number of events, resulting in a substantial computational burden for the E-step. 

\subsubsection{Composite Likelihood Expectation–Maximization Algorithm} 
\label{sec:CLEM}

To alleviate the computational burden imposed by the EM algorithm, this paper proposes a composite likelihood (CL) approach, which facilitates statistical estimation and inference by constructing a pseudo-likelihood as the product of a collection of tractable component likelihoods, with the specific components selected according to the modeling context. Building upon this approach, a CL version EM algorithm, CLEM, that maximizes CL to estimate the model parameter $\boldsymbol{\Theta}$, in the presence of missing label, i.e., latent EP represented by the latency indicators, $I_{m_{s}^{i}}$'s, in Eq. (7). To the best of our knowledge, this work is the first to develop a CLEM algorithm for point process model inference. Specifically, we formulate a CLEM framework for point processes by partitioning the observation window into multiple sub-windows and constructing a CL from block-likelihood contributions evaluated within each sub-window. Unlike conventional CLEM algorithms that typically combine marginal or conditional likelihood components, the proposed method preserves the structural form of the point process likelihood locally within each sub-window based on observed events. This formulation preserves the essential EP mechanism within localized temporal neighborhoods while improving computational tractability by imposing a finite temporal support on the triggering effect. Such an assumption is practically reasonable in autonomous driving contexts, where decision-making primarily relies on most recent data inputs. As new data continuously arrive, earlier errors are progressively corrected or overridden, thereby limiting their long-term propagated influence.

Let the observation window $[0, T)$ be partitioned into $K$ non-overlapping sub-windows of equal length $d = T/K$, where $T$ is the length of observation window. The selection of $K$ will be discussed in detail in Section \ref{subsec: SelectionofK}. For each sub-window $k, k=1,..., K$, we redefine the module-level, stage-level, and system-level error as $\mathbf{t}_{m_s}^k = \{t_{m_s^i} \in \mathbf{t} \mid (k-1)d \leq t_{m_s^i} <kd\}, \quad \mathbf{t}_{s}^k = \cup_{m_s=1}^{M_s}\mathbf{t}_{m_s}^k, \text{ and } \mathbf{t}^k = \cup_{s=1}^{S}\mathbf{t}_{s}^k$, respectively. The composite log-likelihood is then defined as
\begin{equation}
\label{eq:logcomplikelihood}
    \ell_{CL}(\boldsymbol{\Theta}\mid \mathbf{t}) = \sum_{k=1}^K \ell(\boldsymbol{\Theta}\mid \mathbf{t}^k),
\end{equation}
where each component $\ell(\boldsymbol{\Theta}\mid \mathbf{t}^k)$ is derived analogously to Eqs.~(\ref{eq:Overall log-likelihood})–(\ref{eq:log-likelihood}), but with observation window $[(k-1)d, kd)$ and the restricted event set $\mathbf{t}^k$. Accordingly, the latency indicator for each event $t_{m_s^i}\in\mathbf{t}^k$ can be simplified:  
\begin{equation}
    I_{m_s^i}^k =
\begin{cases}
0, & \text{if $t_{m_s^i}$ is a primary error}, \\
j, & \text{if $t_{m_s^i}$ is triggered by upstream error $t_{m_{s-1}^j}\in \mathbf{t}_{s-1}^k$},
\end{cases}
\end{equation} 

\noindent where, $t_{m_{s-1}^j} \in \mathbf{t}_{s-1}^k$ rather than $t_{m_{s-1}^j} \in \mathbf{t}_{s-1}$ as in Eq.~(\ref{eq:Latency Indicator}), implying that only EP scenarios within each sub-window are considered. In other words, we do not need to track the entire event history of stage $s-1$. In this setting, let $\mathbf{I}_{CL}^k=\{I_{m_s^i}^k: s=1, \cdots, S; m_s = 1, \cdots, M_s; i=1,\cdots, n_{m_s}^k\}$ denote the collection of latency indicators associated with event set $\mathbf{t}^k$ where $n_{m_s}^k$ is the number of errors in module $m_s$ occurred in interval $k$. The collection of latency indicators aross all sub-windows is then defined as  $\mathbf{I}_{CL} = \cup_{k=1}^K\mathbf{I}_{CL}^k$. Evaluating the distribution of each latency indicator $I_{m_s^i}^k$ requires only computing the the conditional probabilities: $p_{m_s^i}^{(0)} = \Pr(I_{m_s^i}^k = 0)$, representing the probability that $t_{m_s^i}$ is a primary error, and $p_{m_s^i}^{(j)} = \Pr(I_{m_s^i}^k = j)$, representing the error $t_{m_s^i}$ is a propagated error triggered by the upstream error $t_{m_{s-1}^j}\in \mathbf{t}_{s-1}^k$.

With $\ell_{CL}(\boldsymbol{\Theta}\mid \mathbf{t})$ defined in Eq.~\eqref{eq:logcomplikelihood},  
our objective is to develop a CLEM algorithm that can produce the maximum CL estimation of the model parameter $\boldsymbol{\Theta}$ in the presence of missing data, $\mathbf{I}_{CL}$. Suppose the CLEM algorithm has completed the $a$th iteration and produced an update $\hat{\boldsymbol{\Theta}}^{(a)}, a=0, 1, 2, \cdots$. At the $(a+1)^\text{th}$ iteration, the CL-E step requires calculating the expected value of complete data $\ell_{CL}(\boldsymbol{\Theta}\mid \mathbf{t})$ with respect to the observed data $\mathbf{t}$, the conditional distribution of missing data $\mathbf{I}_{CL}$, and the current estimates $\hat{\boldsymbol{\Theta}}^{(a)}$, i.e.,
\begin{equation}
\begin{aligned}
\label{eq:Qfunction}
\mathcal{Q}\!\left(\boldsymbol{\Theta}\mid \hat{\boldsymbol{\Theta}}^{(a)}\right) 
&= \mathbb{E}_{\mathbf{I}_{CL}\mid \mathbf{t}, \hat{\boldsymbol{\Theta}}^{(a)}}
   \Big[\, \ell_{CL}\!\left(\boldsymbol{\Theta}\mid \mathbf{t}, \mathbf{I}_{CL}\right) \,\Big] = \sum_{k=1}^K \mathbb{E}_{\mathbf{I}_{CL}^k\mid \mathbf{t}^k, \hat{\boldsymbol{\Theta}}^{(a)}}
   \Big[\, \ell\!\left(\boldsymbol{\Theta}\mid \mathbf{t}^k, \mathbf{I}_{CL}^k\right) \,\Big]. 
\end{aligned}
\end{equation}

It is worth noting that, in evaluating the $\mathcal{Q}$ function in Eq.~(\ref{eq:Qfunction}), the proposed CLEM only requires computing likelihood contributions based on subset-specific data $\mathbf{t}^k$,  i.e., $\ell(\boldsymbol{\Theta}\mid \mathbf{t}^k, \mathbf{I}_{CL}^k)$ and then aggregating them across all sub-windows. In contrast, the conventional EM algorithm requires evaluating the likelihood using the full dataset $\mathbf{t}$, i.e., $\ell(\boldsymbol{\Theta}\mid\mathbf{t}, \mathbf{I})$, which leads to substantially higher computational complexity.

The proposed CLEM algorithm iterates the CL-E and CL-M steps until convergence.
\begin{itemize}
    \item \textbf{CL-E Step:} Given the current update $\hat{\boldsymbol{\Theta}}^{(a)}$, obtaining the expected CL function $\mathcal{Q}$ defined in Eq. (\ref{eq:Qfunction}).
    \item \textbf{CL-M Step:} Maximize $\mathcal{Q}\!\left(\boldsymbol{\Theta}\mid \hat{\boldsymbol{\Theta}}^{(a)}\right)$ with respect to $\boldsymbol{\Theta}$ to produce an updated  $\boldsymbol{\Theta}^{(a+1)}$.
\end{itemize}
Details of the $\mathcal{Q}$-function calculation in the CL-E step and the parameter updates for $\boldsymbol{\Theta}$ in the $(a+1)^\text{th}$ iteration of the CL-M step are provided in Appendix~\ref{app:I}.

In the proposed CLEM algorithm, the observation window is partitioned into $K$ sub-windows, and a CL is constructed by considering only the EP effects within each sub-window. The choice of $K$ represents a trade-off between statistical accuracy and computational efficiency. If $K$ is too large, each sub-window becomes temporally narrow, causing many cross-sub-window EP effects to be ignored and potentially leading to biased parameter estimation. For example, missing cross-sub-window EP effects tends to underestimate the EP propagation strength parameter $\alpha_{m_s,m_{s-1}}$ in Eq. {\ref{eq: propagated intensity}}, as some true triggering contributions are omitted. On the other hand, if $K$ is too small, the CL formulation approaches the full-likelihood formulation, thereby diminishing the computational advantage of CLEM. 

\subsubsection{Selection of $K$}
\label{subsec: SelectionofK}

To systematically determine an optimal $K$ value, we propose a stepwise Friedman testing framework.  The Friedman test is a nonparametric group hypothesis test commonly used to test whether the median performance of multiple algorithms or treatments evaluated on the same datasets is equal. It does not assume normality or homoscedasticity of the data \citep{sheldon1996use}. Instead, it relies on within-replication rankings of the performance measures, making it particularly suitable for comparing the performance of the proposed CLEM algorithms with different choices of $K$ \citep{xia2025penalized}. Specifically, we consider a candidate set $\mathcal{K}=\{K_1=1 < K_2 < \cdots < K_M\}$ and construct the corresponding partition set $\mathcal{M}=\{\text{EM},\text{CLEM}(K_2),\ldots,\text{CLEM}(K_M)\}$, where CLEM reduces to EM when $K=1$. Each method is evaluated using a predefined performance metric $e_{\{.,.\}}$ on the same dataset across $R$ independent replications. In this paper, we use the relative root mean square error discussed in Section \ref{subsubsec:Estimation Accuracy} as a performance metric. Let $b=1,2,...$ denote the step index corresponding to candidate value $K_{b+1}$. Starting from $b=1$, we gradually expand the comparison set  $\mathcal{M}_b =\{\text{EM},\text{CLEM}(K_2),\ldots,\text{CLEM}(K_{b+1})\}$ to include CLEM with increasingly larger $K_{b+1}$. At each step $b$, a Friedman test is performed to examine whether the median performances of the methods in the current comparison set remain statistically indistinguishable. If the null hypothesis of equal medians is not rejected, the comparison set is expanded further. Once the null hypothesis is rejected, the procedure stops, and the optimal $K^\ast$ is selected as the largest $K_{b+1}$ whose performance remains statistically comparable to that of EM. The complete stepwise Friedman testing procedure is summarized in Algorithm~\ref{alg:stepwise_friedman}.

\begin{algorithm}[t]
\caption{Stepwise Friedman Test for Selecting Optimal $K^*$}
\label{alg:stepwise_friedman}
\DontPrintSemicolon
\KwIn{Candidate set $\mathcal{K}=\{K_1=1<K_2<\cdots<K_M\}$; method sets $\mathcal{M}=\{\text{EM},\text{CLEM}(K_2),\ldots,\text{CLEM}(K_M)\}$;
the number of repeated estimation runs conducted for each method, i.e., $R$; performance metric $e_{(\cdot, \cdot)}$; significance level $\alpha$.
}
\KwOut{Selected optimal $K^\ast$ such that the median estimation performances of CLEM($K^*$) is not significantly different from the median estimation performance of EM algorithm.}
\BlankLine
\textbf{Methods:} \For {\text{method} $m \in \mathcal{M}$}
{
Run method $m$ on the same data for $R$ replications and record performance matrix
$e_{\{.,.\}}$\;
}
\BlankLine

\textbf{Stepwise testing.}

Initialize $b\leftarrow1$, $K^\ast\leftarrow K_1$\;

\While{true}{

Construct comparison set  
$\mathcal{M}_b=\{\text{EM},\text{CLEM}(K_1),\ldots,\text{CLEM}(K_{b+1})\}$\;


Perform a Friedman test at level $\alpha$ for  
\[
H_0^{(b)}:\ \text{the median estimation performances of all methods in } \mathcal{M}_b \text{ are equal}
\]
\[
H_1^{(b)}:\ \text{not all medians are equal}.
\]
\eIf{$H_0^{(b)}$ is rejected}{
\textbf{break}\tcp*{stop immediately}
}{
$b \leftarrow b+1$; $K^\ast \leftarrow K_{b+1}$\; \tcp*{Expand comparison set}
}

\If{$b=M$}{
\textbf{break}
}
}

\Return{$K^\ast$}\;
\end{algorithm}

\subsubsection{Computational Complexity and Statistical Property}
\label{subsubsec:Computational Analysis and Theoretical Property}

While the stepwise Friedman test provides a data-driven strategy for selecting $K$, this choice introduces a trade-off between computational efficiency and statistical accuracy. To fully understand this trade-off, this paper formally examines both the computational complexity and the statistical properties of CLEM, with particular emphasis on how the value of $K$ influences the CLEM performance. 

    \begin{figure}
        \centering
    \includegraphics[width=1\linewidth]{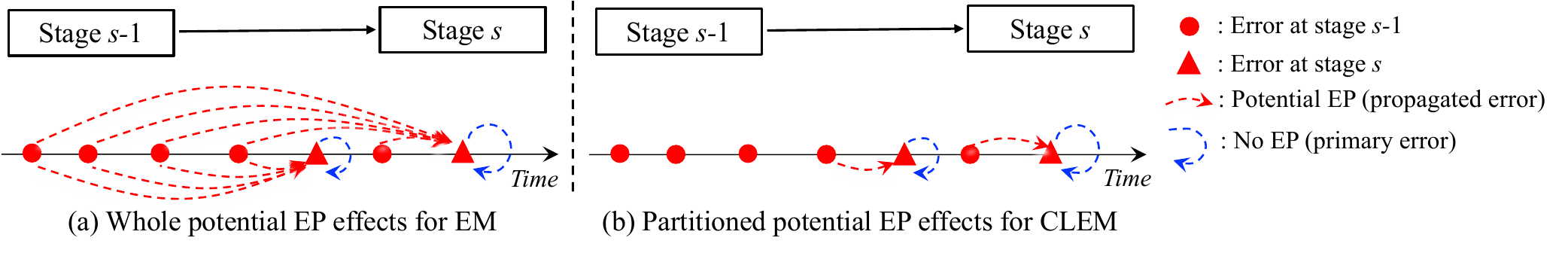}
        \caption{Comparing the Computational Loads of EM and that of CLEM.} 
        \label{fig:EMCLEM}
    \end{figure}

\noindent\textbf{Computational Complexity Analysis.} In an EM algorithm, each E-step requires calculating probabilities of the form $p_{m_s^i}^{(0)}$ and $p_{m_s^i}^{(j)}$'s. This involves tracking the entire event history $t_{m_{s-1}^j} \in \mathbf{t}_s$ for every error $t_{m_s^i} \in \mathbf{t}$, leading to a computational complexity of $\mathcal{O}(N^2)$, where $N = |\mathbf{t}|$ and $|\cdot|$ denotes the cardinality operator, indicating the number of elements in a set. Such quadratic complexity becomes prohibitive when $N$ is large. However, the proposed CLEM algorithm restricts EP to occur only within sub-windows of length $d$. Each CL-E step requires tracking only the partial history $t_{m_{s-1}^j} \in \mathbf{t}^k_{s-1}$ for each error $t_{m_s^i} \in \mathbf{t}^k$, with $k \in [K]$. This strategy reduces the complexity to $\mathcal{O}\!\left(\sum_{k=1}^K N_{k}^2\right)$, where $N_{k} =|\mathbf{t}^k|$, yielding a significant savings when $\sup_k |\mathbf{t}^k| \ll |\mathbf{t}|$. A simplified example is shown in Fig.~\ref{fig:EMCLEM} to illustrate the computational cost improvements of the EM and the CLEM algorithms. In this example, there are five (5) errors at stage $s-1$ (i.e., red dots) and two (2) errors at stage $s$ (i.e., red triangles). Fig.~\ref{fig:EMCLEM} (a) shows the computational cost of the EM algorithm, where every historical error at stage $s-1$ has a probability to propagate to stage $s$, as indicated by the dashed red lines, resulting in eight possible EP scenarios, all of which must be evaluated during the E-step of the EM algorithm. In contrast,  Fig. \ref{fig:EMCLEM} (b) presents the computational cost of the CLEM algorithm, where the observation window is divided into three sub-windows. Only EP scenarios within each sub-window are considered, while cross-sub-window scenarios are excluded. As a result, the number of potential EP scenarios is reduced to two, meaning that the CL-E step needs to evaluate only these two scenarios.  This reduction directly lowers the number of likelihood evaluations required and, thus, substantially decreases the overall computational burden. This simplified example uses only a few errors for illustration. However, the computational savings become more significant in practice when the number of events is large.

\noindent\textbf{Ascent Property.}  
Although the proposed CLEM algorithm employs a composite log-likelihood to approximate the full log-likelihood, it preserves the desirable ascent property. In other words, the composite log-likelihood is guaranteed to be non-decreasing at each CLEM iteration. This monotonicity is important for ensuring numerical stability and supporting convergence of the iterative estimation procedure. Formally, given the observed data $\mathbf{t}$, the composite log-likelihood $\ell_{CL}(\boldsymbol{\Theta}\mid\mathbf{t})$ defined in (\ref{eq:logcomplikelihood}) and the sequence of CLEM parameter estimators $\{\hat{\boldsymbol{\Theta}}^{(a)}\}_{a=0,1,...}$ satisfy
\begin{equation}
\ell_{CL}(\hat{\boldsymbol{\Theta}}^{(a)} \mid \mathbf{t})
\le
\ell_{CL}(\hat{\boldsymbol{\Theta}}^{(a+1)} \mid \mathbf{t}),
\label{eq:theorem1}
\end{equation}
with equality holding if and only if $
\mathcal{Q}(\hat{\boldsymbol{\Theta}}^{(a+1)} \mid \mathbf{t}, \mathbf{I}_{CL}, \hat{\boldsymbol{\Theta}}^{(a)})
=
\mathcal{Q}(\hat{\boldsymbol{\Theta}}^{(a)} \mid \mathbf{t}, \mathbf{I}_{CL}, \hat{\boldsymbol{\Theta}}^{(a)})$. 

\noindent\textbf{Statistical consistency.}
To study the asymptotic behavior of the proposed CLEM estimator, consider the composite score function associated with the composite log-likelihood in (\ref{eq:logcomplikelihood}),
\begin{equation}
U_{K,d}(\boldsymbol{\Theta})
= \sum_{k=1}^K U_{k,d}(\boldsymbol{\Theta}),
\end{equation}
where
\begin{equation}
U_{k,d}(\boldsymbol{\Theta})
= \frac{\partial}{\partial \boldsymbol{\Theta}}
\log f(\mathbf{t}^k \mid \boldsymbol{\Theta}),
\end{equation}
$f(\mathbf{t}^k \mid \boldsymbol{\Theta})$ denotes the likelihood contribution from the $k$th sub-window, $d$ represents the sub-window length. Let $\hat{\boldsymbol{\Theta}}_{K,d}$ denote the CLEM estimator defined as the solution to $U_{K,d}(\boldsymbol{\Theta})=0$ and $\boldsymbol{\Theta}_0$ denote the true parameter vector. Assume that the parameter space of $\boldsymbol{\Theta}$ is compact, and the following conditions are
satisfied,
\begin{enumerate}
     \item [(i)] The expected composite score satisfies $\mathbb{E}[U_{K,d}(\boldsymbol{\Theta})]=0$ if and only if $\boldsymbol{\Theta}=\boldsymbol{\Theta}_d$, where $\boldsymbol{\Theta}_d$ denotes the pseudo-true parameter associated with the composite likelihood approximation;
    \item [(ii)] There exists a nonnegative function $\tau(\cdot)$ such that

\begin{equation}
    \|\frac{f^{(1)}(\mathbf{t}^k \mid \boldsymbol{\Theta})}{f(\mathbf{t}^k \mid \boldsymbol{\Theta})}\| < \tau(n^k),
\end{equation}
\begin{equation}
    \|\frac{f^{(2)}(\mathbf{t}^k \mid \boldsymbol{\Theta})}{f(\mathbf{t}^k \mid \boldsymbol{\Theta})}\| < \tau(n^k),
\end{equation}
\noindent where $\mathbb{E}[\tau(n^k)^2] < \infty$ and $n^k=|\mathbf{t}^k|$ denotes the number of observed events in the $k$th sub-window. 
\end{enumerate}
Then, the CLEM estimator $\hat{\boldsymbol{\Theta}}_{K,d}$ converges in probability to $\boldsymbol{\Theta}_d$ as the number of sub-windows $K \to \infty$. Furthermore, if $\mathbb{E}[U_{K,d}(\boldsymbol{\Theta})] \to 0$ as sub-window length $d \to \infty$ only when $\boldsymbol{\Theta} = \boldsymbol{\Theta}_0$, then $\boldsymbol{\Theta}_d \to \boldsymbol{\Theta}_0$ as $d \to \infty$. The first result implies that, for a fixed $d$, increasing $K$ yields convergence of the estimator to a pseudo-true parameter $\boldsymbol{\Theta}_d$, which maximizes the composite likelihood. In general, $\boldsymbol{\Theta}_d \neq \boldsymbol{\Theta}_0$ because restricting EP to localized temporal neighborhoods introduces an approximation to the full likelihood. The second result indicates that as the sub-window length $d$ increases, the locality constraint weakens and the composite likelihood more closely approximates the full likelihood. Consequently, $\boldsymbol{\Theta}_d \to \boldsymbol{\Theta}_0$. These results show that the proposed CLEM algorithm is statistically consistent when the effective sample size increases, that is, as the number of sub-windows $K \to \infty$, the sub-window length $d \to \infty$, and consequently the total observation horizon $T = K\cdot d \to \infty$. The proof  of ascent property and statistical consistency are provided in the Appendix \ref{app:II}.

\subsection{Model Evaluation}
\label{subsubsec:Model evaluation}
The performance of the proposed model is evaluated in terms of estimation accuracy, prediction accuracy, and computational efficiency. 
 
\noindent \textbf{Estimation accuracy.}  
    The relative root mean square error (RRMSE) \citep{min2022reliability} is used to measure the accuracy of parameter estimation. It is defined as
\begin{equation}
    RRMSE = \sqrt{\frac{1}{P}\sum_{p=1}^{P}\left(\frac{\theta_p-\hat{\theta}_p}{\theta_p}\right)^2},
    \label{eq: RRMSE_r}
\end{equation}
where $\theta_p \in \boldsymbol{\Theta}$ and $\hat{\theta}_p \in \hat{\boldsymbol{\Theta}}$ denote the true and estimated values of the $p$th parameter, respectively, $p=1, ..., P$, and $P$ is the total number of parameters. The mean RRMSE (MRRMSE) across $R$ replications is then computed as 
\begin{equation}
    MRRMSE = \frac{1}{R}\sum_{r=1}^{R}RRMSE_r.
    \label{eq: MRRMSE}
\end{equation}

\noindent \textbf{Prediction Accuracy.}  
    The mean absolute error (MAE) is adopted to assess prediction accuracy. Suppose error events occurring in module $m_s$ are observed up to time $\tau$. Using these events, parameters can be estimated via the proposed method in Section \ref{sec: Model Estimation}. Substituting the estimates into Eqs.~(\ref{eq:Total intensity})--(\ref{eq: propagated intensity}) yields the predictive intensity $\hat{\lambda}_{m_s}(t)$. Based on this, the predicted number of error events during a future interval $[\tau, \tau+\Delta\tau)$ is given by \citep{van2012estimation}:
\begin{equation}
    \hat{N}_{m_s|[\tau, \tau+\Delta\tau)} = \int_{\tau}^{\tau+\Delta\tau} \hat{\lambda}_{m_s}(t) \, dt.
    \label{eq:pred_N} 
\end{equation}

Let $N_{m_s|[\tau, \tau+\Delta\tau)}$ denote the actual number of error events observed in the same interval. For replication $r$ ($r=1,\dots,R$), denote the predicted and actual number of error events as $\hat{N}^r_{m_s|[\tau, \tau+\Delta\tau)}$ and $N^r_{m_s|[\tau, \tau+\Delta\tau)}$, respectively. The MAE is then defined as
\begin{equation}
    MAE_{m_s|[\tau, \tau+\Delta\tau)} = \frac{1}{R}\sum_{r=1}^R \big|N^r_{m_s|[\tau, \tau+\Delta\tau)} - \hat{N}^r_{m_s|[\tau, \tau+\Delta\tau)}\big|,
    \label{eq:MAE} 
\end{equation}
which evaluates prediction performance in module $m_s$ during $[\tau, \tau+\Delta\tau)$. 

\noindent\textbf{Computational Efficiency.}  
    Computational efficiency is measured by the average computation time across $R$ replications. 
    
\begin{figure}
    \centering
    \includegraphics[width=1\linewidth]{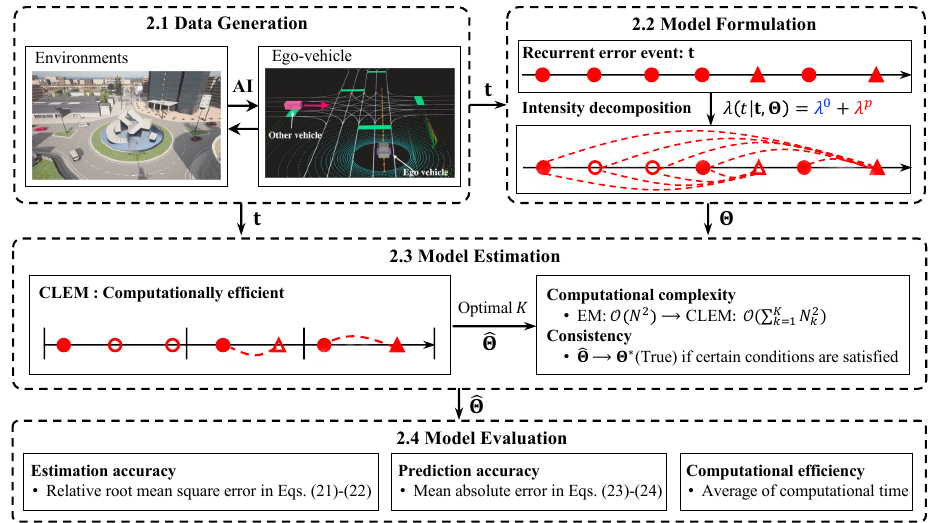}
    \caption{An Illustration of the Proposed Methodological Framework}
    \label{fig: Methodology summary}
\end{figure}

To summarize, the methodology framework is illustrated in Fig. \ref{fig: Methodology summary}.

\section{Case Study}
\label{sec: Case Study}

This section demonstrates the effectiveness and computational efficiency of the proposed CLEM-based EP modeling framework for AI systems reliability analysis. Section~\ref{subsec: Numerical Case Study} presents a numerical simulation showing that, with an appropriate choice of $K$, the proposed CLEM algorithm substantially reduces computational time while maintaining estimation accuracy comparable to that of the standard EM algorithm. Section~\ref{subsec: Physics-based Case Study} reports the results from a physics-based AV simulation, demonstrating that the proposed EP modeling framework improves reliability prediction performance relative to benchmark models.

\subsection{Numerical Case Study}
\label{subsec: Numerical Case Study}
\subsubsection{Experiment Setup} 
\begin{figure}
    \centering
    \includegraphics[width=0.6\linewidth]{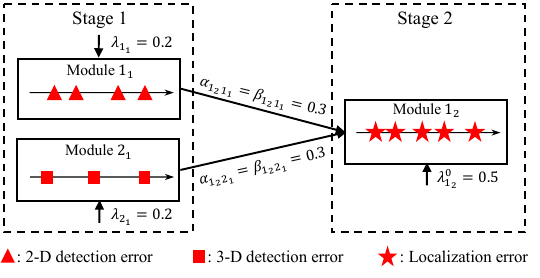}
    \caption{Numerical Simulation Scenario}
    \label{fig:simulatedscenarios}
\end{figure}

In this case study, we simulate error event data from two consecutive stages (i.e., Stage $1$ and Stage $2$), using the thinning algorithm proposed in \citep{lewis1979simulation}. Specifically, we consider two modules at Stage $1$ (i.e., $m_{1_1}$ and $m_{2_1}$) and one module at Stage $2$ (i.e., $m_{1_2}$), as shown in Figure \ref{fig:simulatedscenarios}. For $m_{1_1}$ and $m_{2_1}$ at stage 1, the error events are simulated from a univariate homogeneous Poisson process with intensity $\lambda_{1_1} = \lambda_{2_1} = 0.2$. For the $m_{1_2}$ at stage $2$, the error events are simulated with intensity,
\begin{equation}
 \label{eq:exp_intensity}
    \lambda_{1_2}(t)
    = \lambda_{1_2}^0 +  \sum\nolimits_{m=1}^{2}\sum_{j: t_{m_{1}^j}<t}  \alpha_{1_2, m_{1}} \cdot \beta_{1_2, m_{1}} \cdot \exp\{-\beta_{1_2, m_{1}} \cdot (t-t_{m_{1}^j})\},
\end{equation}
where we set the parameters as follows: $\lambda_{1_2}^0 = 0.5$, and $\alpha_{1_2, 1_{1}} = \alpha_{1_2, 2_{1}} = \beta_{1_2, 1_{1}} = \beta_{1_2, 2_{1}} = 0.3$. These values are chosen to roughly align with the error rates reported for each module \citep{pan2024reliability}. To simulate datasets of varying sizes, we set the simulation window length 
$T$ to 500, 1000, 2500, and 5000. For each parameter configuration, we repeatedly generated three types of error events from three distinct modules for 100 times (i.e., $R=100$). Model estimation was performed using both the EM and the CLEM algorithms, with the number of sub-windows $K$ set to 2, 5, 10, 20, 50, 100, 250, and 500. 

\subsubsection{Result - Estimation Accuracy} 
\label{subsubsec:Estimation Accuracy}

Table \ref{result_estimationacc} presents the means and standard deviations of 100 repeated parameter estimates using the EM algorithm and the CLEM algorithm with different choices of $K$. The results show that the mean of CLEM estimates with $K=2,5,10,20,50, \text{and } 100$, as well as those from the EM algorithm, closely align with the true parameter values. However, noticeable bias emerges when $K =250, \text{and } 500$, which indicates that excessively large values of $K$ may reduce estimation accuracy. This reduction occurs because, under a fixed simulation time window $T$,  increasing $K$ divides the window into many smaller sub-windows. The CLEM algorithm then approximates the full likelihood using CLs constructed within each small sub-windows, potentially ignoring important temporal dependencies and resulting in information loss. The results shown in Table \ref{result_estimationacc} are based on a simulation an overall AI system operation time window of $T=5,000$. 
\begin{table*} 
\small
\centering
\begin{threeparttable}
\caption{Parameter Estimation Result Based on EM and CLEM Algorithms (T=5,000)}
\label{result_estimationacc}
\begin{tabular}{lllllll}
\hline
\hline
\textbf{Method} & ${K}$ & $\hat{\lambda}_{m_s}^0$ & $\hat{\alpha}_{{m_s}, 1_{s-1}}$ & $\hat{\alpha}_{{m_s}, 2_{s-1}}$ & $\hat{\beta}_{{m_s}, {2_{s-1}}}$ & $\hat{\beta}_{{m_s}, 2_{s-1}}$ \\
 \hline
 True & - & 0.5 & 0.3 & 0.3 & 0.3 & 0.3\\
\hline
EM & - & 0.481$^{\S}$ (0.015)\textsuperscript{\textdagger} & 0.295 (0.047) & 0.294 (0.037) & 0.287 (0.040) & 0.289 (0.035) \\
\hline
\multirow{8}{*}{CLEM} 
& 2   & 0.480 (0.015) & 0.295 (0.044) & 0.295 (0.039) & 0.286 (0.041) & 0.289 (0.036) \\
& 5   & 0.480 (0.015) & 0.297 (0.045) & 0.298 (0.042) & 0.284 (0.037) & 0.287 (0.035) \\
& 10  & 0.480 (0.016) & 0.298 (0.042) & 0.298 (0.039) & 0.284 (0.039) & 0.286 (0.034) \\
& 20  & 0.482 (0.015) & 0.295 (0.045) & 0.295 (0.041) & 0.285 (0.039) & 0.288 (0.032) \\
& 50  & 0.487 (0.015) & 0.287 (0.045) & 0.288 (0.043) & 0.289 (0.039) & 0.289 (0.034) \\
& 100 & 0.494 (0.014) & 0.281 (0.041) & 0.282 (0.039) & 0.290 (0.040) & 0.291 (0.032) \\
& 250 & 0.512 (0.013) & 0.261 (0.043) & 0.260 (0.038) & 0.290 (0.041) & 0.290 (0.038) \\
& 500 & 0.533 (0.013) & 0.251 (0.041) & 0.250 (0.039) & 0.261 (0.039) & 0.262 (0.037) \\
\hline
\hline
\end{tabular}
\begin{tablenotes}
\footnotesize
\item  $^{\S}$: Mean value across 100 estimates.
\item \textsuperscript{\textdagger}: Values in parentheses represent the standard deviation of the 100 estimates.
\end{tablenotes}
\end{threeparttable}
\end{table*}

\begin{figure}
    \centering    \includegraphics[width=0.6\linewidth]{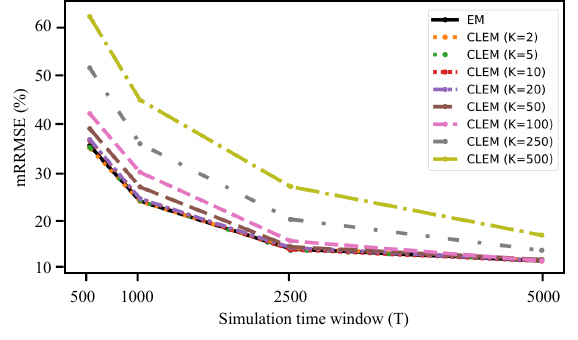}
    \caption{MRRMSE Values of EM and CLEM Algorithms}
    \label{fig:MRRMSE}
\end{figure}

Table \ref{result_estimationacc} provides the estimation results of each individual parameter. To evaluate the overall estimation accuracy of EM and CLEM algorithms, we use the MRRMSE metric defined in Eqs.(\ref{eq: RRMSE_r})-(\ref{eq: MRRMSE}). Fig. \ref{fig:MRRMSE} presents the MRRMSE values for the EM and the CLEM algorithms with different choices of $K$ across different lengths of the simulation time window $T$. The results show that when $K = 2, 5, 10$, and $20$, the MRRMSE values of the CLEM algorithm are comparable to those of the EM algorithm across all values of $T$, indicating that dividing $T$ into a small number of sub-windows in CLEM does not compromise estimation accuracy. In addition, as $T$ increases, the CLEM algorithm can tolerate a larger number of sub-windows without a loss in accuracy. For instance, when $T = 5,000$, the MRRMSE is still comparable with the EM algorithm even when $K$ increases to 50 or 100. However, when $K$ becomes excessively large (e.g., $K = 250$ or $500$), the MRRMSE of the CLEM algorithm increases substantially, which is due to the information loss incurred when the full likelihood is approximated by CLs over a large number of small sub-windows.

\subsubsection{Result - Computational Efficiency}\label{subsubsec:Computational Efficiency}
\begin{figure}
    \centering
\includegraphics[width=0.6\linewidth]{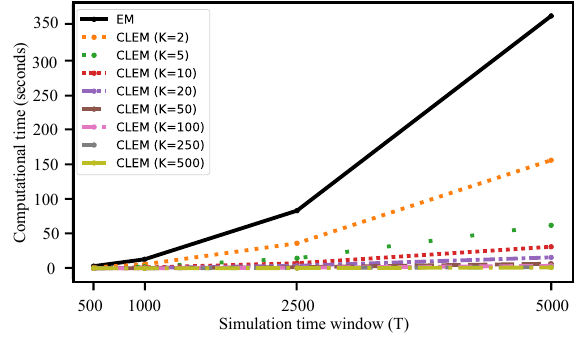}
    \caption{Computational Time of EM and CLEM Algorithms}
    \label{fig:runningtime}
\end{figure}

Fig. \ref{fig:runningtime} presents the computational time required by the EM and the CLEM algorithms for different $K$ across different lengths of simulation time window $T$. There are several interesting findings. First, as expected, a larger $T$ results in longer computational times for both methods. This is because a larger $T$ leads to a larger dataset size, increasing the computational burden. This result highlights the importance of developing computationally efficient algorithms for large-scale data settings and longer AI system operations. Second, compared to the EM algorithm, the CLEM algorithm demonstrates a significant reduction in computation time, especially as $K$ and $T$ increase. This improvement arises because, by dividing the data into smaller segments, the CLEM algorithm reduces the computational load within each sub-window. Third, when the dataset size is small (e.g., $T = 500$), the computational advantage of the CLEM over the EM becomes less significant. 

\subsubsection{Optimal $K$ Selection}
\begin{figure} 
    \centering
    \includegraphics[width=1\linewidth]{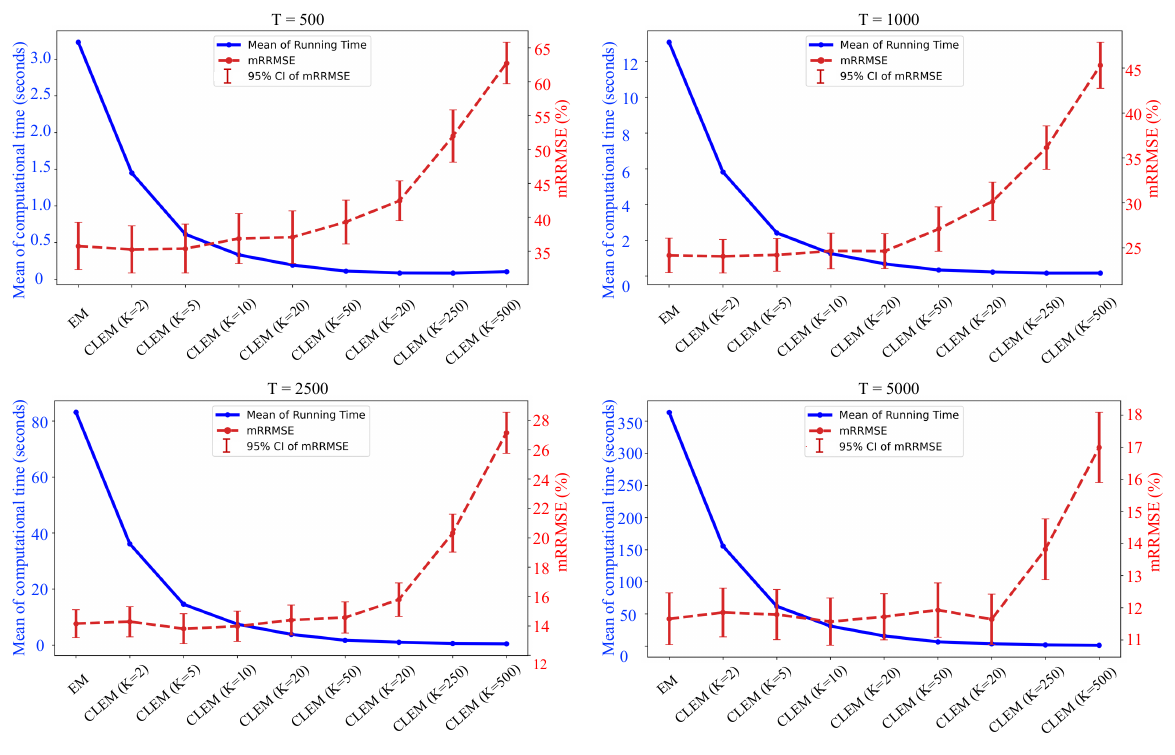}
    \vspace{-2em}
    \caption{Comparison of MRRMSE and Computation Time}
    \label{fig:RRMSE}
\end{figure}

The results in Sections \ref{subsubsec:Estimation Accuracy} and \ref{subsubsec:Computational Efficiency} demonstrate that the choice of $K$ influences both estimation accuracy and computational efficiency of the CLEM algorithm. To further illustrate this trade-off, Fig.~\ref{fig:RRMSE} presents the MRRMSE values and computational times of the EM and the CLEM algorithms with varying $K$. In each subplot, the red dashed line indicates the MRRMSE, with the red interval representing the 95\% confidence interval (CI) of RRMSE values across 100 replications. The blue solid line represents the corresponding mean computational time. This figure clearly illustrates that increasing $K$ yields substantial reductions in computational time, but at the expense of increased MRRMSE values. These findings highlight the importance of carefully selecting an appropriate $K$ to avoid compromising estimation accuracy for the sake of computational efficiency.

The stepwise Friedman test introduced in Section~\ref{subsec: SelectionofK} is employed to determine the optimal $K^*$ for the CLEM algorithm. In this case study, the candidate set is $\mathcal{K} = \{2, 5, 10, 20, 50, \\ 100, 250, 500\}$, and the corresponding model set is $\mathcal{M} = \{\text{EM}, \text{CLEM}(2), \text{CLEM}(5), \text{CLEM}(10), \\ \text{CLEM}(20), \text{CLEM}(50), \text{CLEM}(100), \text{CLEM}(250), \text{CLEM}(500)\}$. The RRMSE is used as the evaluation metric. The objective is to select the largest $K^* \in \mathcal{K}$ such that the median RRMSE of $\text{CLEM}(K^*)$, denoted $M_{\text{CLEM}(K^*)}$, is not significantly different from the median RRMSE of EM, denoted $M_{\text{EM}}$. We begin with step 1 and the comparison set is $\mathcal{M}_1 = \{\text{EM}, \text{CLEM}(2), \text{CLEM}(5)\}$. The hypothesis is formulated as:
\[
\begin{aligned}
H_0^{(1)}:&\; M_{EM} = M_{\text{CLEM}(2)} = M_{\text{CLEM}(5)}; \\
H_1^{(1)}:&\; \text{Not all medians are equal}.
\end{aligned}
\]

The results, shown in Table~\ref{tab:Friedman test}, fail to reject $H_0^{(1)}$ with $p = 0.181$, indicating no significant difference among $M_{\text{EM}}$, $M_{\text{CLEM}(2)}$, and $M_{\text{CLEM}(5)}$. In step 2, the comparison set is extended to $\mathcal{M}_2 = \{\text{EM}, \text{CLEM}(2), \text{CLEM}(5), \text{CLEM}(10)\}$. Again, the null hypothesis $H_0^{(2)}$ is not rejected with $p = 0.260$. However, beginning with step 3, which includes CLEM(20) in the comparison set, the null hypothesis $H_0^{(3)}$ is rejected with $p = 0.026$, indicating that CLEM(20) introduces significant difference among $M_{\text{EM}}$, $M_{\text{CLEM}(2)}$, \text{CLEM}(5), \text{CLEM}(10), and $M_{\text{CLEM}(20)}$. Based on these results, we select $K^* = 10$ as the optimal number of sub-windows. Table~\ref{tab:Friedman test} reports the stepwise Friedman test results for $T = 500$, while additional results for $T = 1000$, $2500$, and $5000$ are provided in the Appendix \ref{app:III}. Using the same selection criterion, we obtain the following optimal choices: $K^* = 20$ for $T = 1000$, $K^* = 50$ for $T = 2500$, and $K^* = 100$ for $T = 5000$. A noteworthy finding is that the optimal sub-window length, defined as $T/K^*$, remains consistently equal to 50 across all examined values of $T$. This consistency suggests that the performance of the CLEM algorithm depends more fundamentally on the sub-window length than on the total simulation time $T$.

\begin{table*}[t]
\small
\centering
\begin{threeparttable}
\caption{Stepwise Friedman Test Results for Selecting the Optimal $K$ ($T=500$)}
\label{tab:Friedman test}
\begin{tabular}{lll}
\hline
\hline
\textbf{Hypothesis} & \textbf{$p$-value} & \textbf{Conclusion}  \\
\hline
$H_0^{(1)}: M_{EM} = M_{CLEM(2)} = M_{CLEM(5)}$ & \multirow{2}{*}{0.181
} & \multirow{2}{*}{No reject $H_0^1$}   \\
$H_1^{(1)}: \text{Not all medians are equal}$ &  &   \\
\hline
$H_0^{(2)}: M_{EM} = M_{CLEM(2)} = M_{CLEM(5)} =  M_{CLEM(10)}$ & \multirow{2}{*}{0.226

} & \multirow{2}{*}{No reject $H_0^2$}   \\
$H_1^{(2)}: \text{Not all medians are equal}$ &  &   \\
\hline
$H_0^{(3)}: M_{EM} = M_{CLEM(2)} = M_{CLEM(5)} =  M_{CLEM(10)} =  M_{CLEM(20)} $ & \multirow{2}{*}{0.026
} & \multirow{2}{*}{Reject $H_0^3$}\textsuperscript{a}   \\
$H_1^{(3)}: \text{Not all medians are equal}$ &  &   \\
\hline
\hline
\end{tabular}
\begin{tablenotes}
\item[a] Rejection decision is made if $p$-value is smaller that 0.05.
\end{tablenotes}
\end{threeparttable}
\end{table*}

\subsection{Physics-based Simulation Case Study}
\label{subsec: Physics-based Case Study}

This case study is enabled by a physics-based simulation platform equipped with an error injector (EI). The EI provides the capability to inject errors into any functional module $m_s$ at user-defined timestamps $t^{err}_{m_s}$, with each injection occurring under a specified probability $p_{t^{err}_{m_s}}$.

\subsubsection{Experiment Setup}
\label{Experiment setup}

\begin{figure}[htb]
{
\centering
\includegraphics[width=0.55\columnwidth]{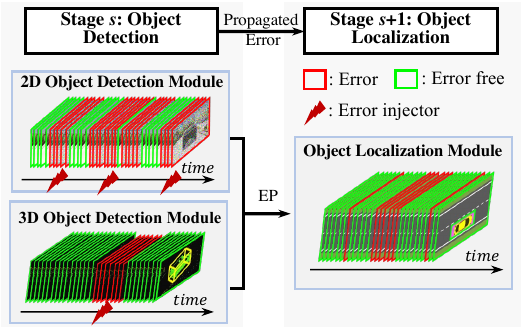}
\caption{Physics-Based Simulation Scenario} 
\label{fig: EI}
}
\end{figure}

This case study investigates EP between the object detection stage and the object localization stage. As illustrated in Fig. \ref{fig: EI}, the object detection stage consists of two modules, i.e., the 2-D detection module and the 3-D detection module, while the object localization stage is composed of a single module, the object localization module. In this study, errors are injected into both modules of the object detection stage, and their subsequent propagation into the object localization stage is examined.

We consider two EI settings, as defined in Section \ref{subsec:Data Generation} and summarized in Table~\ref{tab:error injection percentage}. Setting I consists of four scenarios, in which errors are continuously injected over the entire simulation interval, i.e., $t^{err}_{m_s} \in [0, 200)$, with varying injection probabilities. These scenarios are designed to emulate the performance of the 2-D and 3-D detection modules under persistent weather conditions: clear (Sce. 1), snowy (Sce. 2), rainy (Sce. 3), and foggy (Sce. 4). Setting II, by contrast, introduces errors intermittently, with injections given only during the time intervals $[50, 100)$ and $[150, 200)$. This setting reflects intermittent driving conditions, where errors occur sporadically rather than persistently. Corresponding injection probabilities are applied to simulate the performance of the 2-D and 3-D detection modules under intermittent snow (Sce. 5), rain (Sce. 6), and fog (Sce. 7). The EI probabilities, $p_{t^{err}_{m_s}}$, listed in Table \ref{tab:error injection percentage}, are chosen based on empirical findings reported in \citep{hassaballah2020vehicle, kilic2021lidar}. These values approximate the error occurrence rates of 2-D and 3-D detection modules when operating under real-world adverse weather conditions. In each scenario, error-event data are simulated at a sampling frequency of 20 Hz for a total duration of 200 seconds. Three types of error events are collected, i.e., 2-D miss-detection errors, 3-D miss-detection errors, and miss-localization errors. The simulation is repeated thirty times under each scenario. 

\begin{table*}[t]
    \small
    \centering
        \caption{Error Injection Probability in Object Detection Stage}
    \label{tab:error injection percentage}
    \begin{tabular}{clllllll}
    \hline
    \hline & \multicolumn{4}{c}{Setting I: Persistent EI }  & \multicolumn{3}{c}{Setting II: Intermittent EI} \\
    \cmidrule(lr){2-5} \cmidrule(lr){6-8} & Sce. 1  & Sce. 2  & Sce. 3  & Sce. 4  & Sce. 5 & Sce. 6 & Sce. 7\\
        \cmidrule(lr){2-5} \cmidrule(lr){6-8} & Clear  & Snow  & Rain  & Foggy  & Snow  & Rain  & Foggy\\ 
    \hline
    2-D detection  &  0.00 &  0.40 &  0.55   & 0.60  &  0.40 &  0.55  & 0.60  \\
    3-D detection  &  0.00  &  0.25 &  0.50   & 0.60  &  0.25 &  0.50  & 0.60   \\
    \hline
    \hline
     \end{tabular}
     \end{table*} 

   \begin{table*}
\centering
\small
\caption{Summary of Comparison Methods}
\label{Benchmarks}
\begin{tabular}{l|l|l|l}
\hline\hline
& Benchmarks & $\lambda_{m_s}(t;\boldsymbol{\theta})$  & Parameters \\
\hline
\multirow{3}{*}{\vspace{-0em} NHPP} & Musa-Okumoto (MO) \citep{musa1984logarithmic} & $\theta_2 (1+\theta_2\theta_1 t)^{-1}$                             & \begin{tabular}[c]{@{}l@{}}$\boldsymbol{\theta}=(\theta_1, \theta_2)^\top$\\ $\theta_1 > 0, \theta_2 > 0$\end{tabular}  \\
\cline{2-4}
& Gompertz \citep{ohishi2009gompertz} & $\theta_1 \theta_2^{t} \theta_3^{\theta^t_2} \log{(\theta_2)} \log(\theta_3)$ & \begin{tabular}[c]{@{}l@{}}$\boldsymbol{\theta}=(\theta_1, \theta_2, \theta_3)^\top$\\ $\theta_1 > 0, 0 < \theta_2, \theta_3  < 1$\end{tabular} \\
\hline
HPP  & Poisson \citep{sahinoglu1992compound} & $\theta_1$ & \begin{tabular}[c]{@{}l@{}}$\boldsymbol{\theta} = (\theta_1)^\top$\\ $\theta_1 > 0$\end{tabular}  \\           
\hline\hline                           
\end{tabular}
\end{table*}

 With the simulated data, we evaluate the prediction accuracy of the proposed model relative to the benchmark reliability models summarized in Table~\ref{Benchmarks}. Specifically, we estimate the intensity function of the proposed model and the benchmarks based on data collected from the first 180 seconds (i.e., $\tau = 180$) and subsequently used the estimated intensity function to predict the number of errors in the following interval $[180, 200)$ (i.e., $\Delta \tau = 20$). This procedure is repeated thirty times (i.e., $R=30$). The mean absolute error (MAE) of each model can be calculated by Eqs. (\ref{eq:pred_N})-(\ref{eq:MAE}).

 \subsubsection{Result - Prediction Accuracy}
 \label{Model prediction}

\begin{figure}[htb]
{
\centering
\includegraphics[width=0.7\columnwidth]{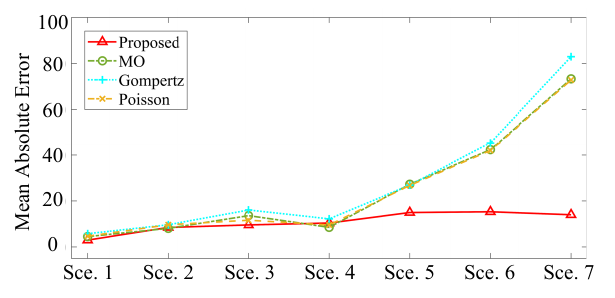}
\caption{Prediction Accuracy Comparison Based on Physics-Based Simulation} 
\label{fig: MAE}
}
\end{figure}

Fig. \ref{fig: MAE} shows the MAEs of the proposed model and benchmarks across seven scenarios in Table \ref{tab:error injection percentage}. Overall, the proposed method (shown as the red line) consistently achieves the lowest MAEs compared with the benchmarks. In Setting I (Sces. 1–4), the proposed model and the benchmarks exhibit comparable prediction accuracy. This similarity can be attributed to the nature of Setting I, where errors are injected continuously throughout the simulation interval. Under these persistent weather conditions, the error occurrence remains relatively stable, allowing the benchmark models to capture the underlying pattern effectively. In contrast, in Setting II (Sces. 5–7), the proposed method demonstrates a significant advantage, with MAEs significantly lower than those of the benchmarks. This improvement indicates that by explicitly considering the EP, the proposed model can more accurately predict error occurrences under \textit{intermittent} EI conditions, where errors arise sporadically and pose greater challenges for conventional benchmark models. This property makes the proposed CLEM-based modeling approach uniquely advantageous in analyzing the reliability of AI systems, which are applied in perception and decision-making in uncontrolled environments.

\section{Conclusion}
\label{sec:Conclusions}

AI system reliability modeling is critical for the development and safe deployment of autonomous technologies. This paper addresses the challenges of data, modeling, and estimation in AI systems reliability analysis, with particular emphasis on error propagation modeling. Leveraging a physics-based simulation platform, we develop a justifiable error injector to systematically generate error event data from AI systems in autonomous vehicles. Based on these simulated data, we propose a generic intensity decomposition framework that explicitly distinguishes between primary errors and propagated errors. To efficiently estimate latent error propagation, we design a CLEM algorithm that is computationally scalable and supported by theoretical guarantees. A numerical case study demonstrates that the CLEM algorithm significantly reduces computation time while maintaining estimation accuracy comparable to the standard EM algorithm. Furthermore, a physics-based simulation case study illustrates that explicitly modeling error propagation enables the proposed framework to achieve higher prediction accuracy than benchmark reliability models.

While this study advances reliability modeling of AI systems by explicitly considering the error propagation, several future research directions remain. First, extending the proposed framework to real-world operational data beyond simulation will be critical for validating its practical applicability and robustness across diverse driving environments. Second, integrating uncertainty quantification and adaptive learning could enhance the model’s ability to update reliability assessments as new data are collected. Finally, extending the methodology to other safety-critical domains, such as healthcare, energy systems, and industrial automation, offers the potential to generalize the framework into a broader class of AI system reliability models.

\appendix
\section{Appendix: Details of CLEM algorithm}
\label{app:I}

\noindent \textbf{CLEM algorithm}  Given an initial parameter estimate $\hat{\boldsymbol{\Theta}}^{(0)}$, the parameter $\boldsymbol{\Theta}$ is updated iteratively by alternating between the CL-E step and the CL-M step. Let $a=0,1,2,\ldots$ denote the iteration index. The $(a+1)^{\text{th}}$ iteration then proceeds according to the following steps:

\begin{itemize}
\item \textbf{CL-E step:} Given $\hat{\boldsymbol{\Theta}}^{(a)}$, the CL-E step requires calculating expected value of complete data $\ell_{CL}(\boldsymbol{\Theta}\mid \mathbf{t})$ with respect to the observed data $\mathbf{t}$, the conditional distribution of missing data $\mathbf{I}_{CL}$, and the current estimates $\hat{\boldsymbol{\Theta}}^{(a)}$, i.e.,
\renewcommand{\theequation}{\thesection.1}
\begin{equation}
\begin{aligned}
\label{eq:CLE-step}
\mathcal{Q}\!\left(\boldsymbol{\Theta}\mid \hat{\boldsymbol{\Theta}}^{(a)}\right) 
&= \mathbb{E}_{\mathbf{I}_{CL}\mid \mathbf{t}, \hat{\boldsymbol{\Theta}}^{(a)}}
   \Big[\, \ell_{CL}\!\left(\boldsymbol{\Theta}\mid \mathbf{t}, \mathbf{I}_{CL}\right) \,\Big] \\[6pt]
&= \sum_{k=1}^K \sum_{s=1}^{S} \sum_{m_s=1}^{M_s} 
   \Bigg\{ \sum_{i=1}^{n_{m_s}^k} 
   \Bigg[ \, 
       p_{m_s^i}^{(0)} \, \log \lambda_{m_s}^0 \\[6pt]
&\quad + \sum_{\substack{j: \, t_{m_{s-1}^j} \in \mathbf{t}_{s-1}^k \\ t_{m_{s-1}^j} < t_{m_s^i}}} 
       p_{m_s^i}^{(j)} \, \log \lambda_{m_s}^p\!\left(t_{m_s^i} - t_{m_{s-1}^j}\right) \\[6pt]
&\quad - \int_{kd-d}^{kd} \lambda_{m_s}(t)\, dt 
   \Bigg] \Bigg\},
\end{aligned}
\end{equation}

\noindent where the probabilities $p_{m_s^i}^{(0)}$ and $p_{m_s^i}^{(j)}$'s explicitly represent if an event $t_{m_s^i}$ is a primary error or a propagated error that is triggered by event $t_{m_{s-1}^j} \in \mathbf{t}_{s-1}^k$. They can be linked to the primary intensity and propagated intensity defined in Eqs.(\ref{eq:Baseline intensity})
-(\ref{eq: propagated intensity}) and defined as,

\renewcommand{\theequation}{\thesection.2}
  \begin{equation}
\label{eq:Probabilities}
  p_{m_s^i}^{(0)} = \frac{\lambda_{m_s}^0}{\lambda_{m_s}(t_{m_s^i})},\,\,\,\,\,\,\,\,\,\,
p_{m_s^i}^{(j)} = \frac{\lambda_{m_s,m_{s-1}}^p(t_{m_s^i} - t_{m_{s-1}^j})}{\lambda_{m_s}(t_{m_s^i})}.
\end{equation}

\item \textbf{CL-M step:} Given probabilities in Eq. (\ref{eq:Probabilities}) and $\hat{\boldsymbol{\Theta}}^{(a)}$, $\hat{\boldsymbol{\Theta}}^{(a+1)}$ can be updated by maximizing $\mathcal{Q}\!\left(\boldsymbol{\Theta}\mid \mathbf{t}, \mathbf{I}_{CL}, \hat{\boldsymbol{\Theta}}^{(a)}\right)$ in Eq.(\ref{eq:CLE-step}), i.e., 
\renewcommand{\theequation}{\thesection.3}
\begin{equation}
\label{eq:clm}
\hat{\boldsymbol{\Theta}}^{(a+1)} = \arg\max_{\boldsymbol{\Theta}\geq0} \, \mathcal{Q}\!\left(\boldsymbol{\Theta}\mid \mathbf{t}, \mathbf{I}_{CL}, \hat{\boldsymbol{\Theta}}^{(a)}\right) .
\end{equation}

Specifically, each parameter is updated by the stochastic gradient descent method as below:
\renewcommand{\theequation}{\thesection.4}
\begin{equation}
    \label{eq:updated lambda_0}
    \hat{\lambda}_{m_s}^{0(a+1)} = \frac{\sum_{k=1}^K\sum_{i=1}^{n_{m_s}^k}p_{m_s^i}}{T},
\end{equation}
\renewcommand{\theequation}{\thesection.5}
\begin{equation}
    \label{eq:updated alpha}
    \hat{\alpha}_{m_s, m_{s-1}}^{(a+1)} = \frac{\sum_{k=1}^K\sum_{j=1}^{n_{m_{s-1}}^k}\sum_{i=1}^{n_{m_s}^k}p_{m_s^i,m_{s-1}^j}}{\sum_{k=1}^K\sum_{j=1}^{n_{m_{s-1}}^k}(1-\exp{(-\beta_{m_s, m_{s-1}}^{(a)}(kd-t_{m_{s-1}^j}))})},
\end{equation}
\renewcommand{\theequation}{\thesection.6}
\begin{equation}
    \label{eq:updated beta}
    \hat{\beta}_{m_s, m_{s-1}}^{(a+1)} = \frac{\sum_{k=1}^K\sum_{j=1}^{n_{m_{s-1}}^k}\sum_{i=1}^{n_{m_s}^k}p_{m_s^i,m_{s-1}^j}}{A+B},
\end{equation}

\noindent where $A = \sum_{k=1}^K\sum_{j=1}^{n_{m_{s-1}}^k}\sum_{i=1}^{n_{m_s}^k}p_{m_s^i,m_{s-1}^j} \cdot (t_{m_s^i} - t_{m_{s-1}^j})$ and $B = \sum_{k=1}^K\sum_{j=1}^{n_{m_{s-1}}^k} \alpha_{m_s, m_{s-1}}^{(a+1)}\cdot
    (kd-t_{m_{s-1}^j})\cdot\exp{(-\beta_{m_s, m_{s-1}}\cdot(kd-t_{m_{s-1}^j}))}$.

\end{itemize}

\section{Appendix: Proof of Ascent Property and Statistical Consistency}
\label{app:II}

\subsection{Proof of Ascent Property}
\label{Proof of Theorem 1}

\noindent\textbf{Ascent Property.} 
Given the observed data $\mathbf{t}$, the composite log-likelihood $\ell_{CL}(\boldsymbol{\Theta} \mid \mathbf{t})$ defined in (\ref{eq:logcomplikelihood}) and the sequence of CLEM estimators $\hat{\boldsymbol{\Theta}}^{(a)},\, a=1,2,\ldots$, satisfy
\renewcommand{\theequation}{\thesubsection.1}
\begin{equation}
    \label{eq: theorem1}
    \ell_{CL}(\hat{\boldsymbol{\Theta}}^{(a-1)} \mid \mathbf{t}) 
    \leq 
    \ell_{CL}(\hat{\boldsymbol{\Theta}}^{(a)} \mid \mathbf{t}),
\end{equation}
with equality if and only if 
$\mathcal{Q}(\hat{\boldsymbol{\Theta}}^{(a)} \mid \mathbf{t}, \mathbf{I}_{CL}, \hat{\boldsymbol{\Theta}}^{(a)}) 
= 
\mathcal{Q}(\hat{\boldsymbol{\Theta}}^{(a-1)} \mid \mathbf{t}, \mathbf{I}_{CL}, \hat{\boldsymbol{\Theta}}^{(a)})$.

Let $\Theta$ denote the parameter space for $\boldsymbol{\Theta}$.  
For each sub-window $k=1,\ldots,K$, define the conditional density of the collection of latency indicators, $\mathbf{I}_{CL}^k = \{I_{m_s^i}^k: s=1,\ldots,S; m_s=1,\ldots,M_s; i=1,\ldots,n_{m_s}^k\}$
given $\mathbf{t}^k$ as 
$\text{Pr}(\mathbf{I}_{CL}^k \mid \mathbf{t}^k, \boldsymbol{\Theta})$.  In the CLEM framework, the expectation step (CL-E step) defines
\renewcommand{\theequation}{\thesubsection.2}

\begin{equation}
\label{eq:E-step}
\begin{aligned}
  \mathcal{Q}(\boldsymbol{\Theta}\mid \boldsymbol{\Theta}^{'})
  &= 
  \sum_{k=1}^K 
  \mathbb{E}_{\mathbf{I}_{CL}^k \mid \mathbf{t}^k, \boldsymbol{\Theta}^{'}} 
  \big[
  \log \text{Pr}(\mathbf{t}^k, \mathbf{I}_{CL}^k \mid \boldsymbol{\Theta})
  \big] \\
  &= 
  \sum_{k=1}^K 
  \sum_{\mathbf{I}_{CL}^k}
  \log \text{Pr}(\mathbf{t}^k, \mathbf{I}_{CL}^k \mid \boldsymbol{\Theta}) 
  \text{Pr}(\mathbf{I}_{CL}^k \mid \mathbf{t}^k, \boldsymbol{\Theta}^{'}).
\end{aligned}
\end{equation}
Analogously, define the composite $H$-function as
\renewcommand{\theequation}{\thesubsection.3}
\begin{equation}
\label{eq:H-function}
H(\boldsymbol{\Theta}\mid\boldsymbol{\Theta}^{'})
=
\sum_{k=1}^K
\sum_{\mathbf{I}_{CL}^k}
\log \text{Pr}(\mathbf{I}_{CL}^k \mid \mathbf{t}^k, \boldsymbol{\Theta})
\text{Pr}(\mathbf{I}_{CL}^k \mid \mathbf{t}^k, \boldsymbol{\Theta}^{'}).
\end{equation}

The following two lemmas are critical in establishing Theorem~1.

\textbf{Lemma 1.}  
For any $(\boldsymbol{\Theta}, \boldsymbol{\Theta}') \in \Theta \times \Theta$,
\[
\mathcal{Q}(\boldsymbol{\Theta} \mid \boldsymbol{\Theta}') - H(\boldsymbol{\Theta} \mid \boldsymbol{\Theta}') 
= 
\ell_{CL}(\boldsymbol{\Theta} \mid \mathbf{t}).
\]

\noindent\emph{Proof.}
\renewcommand{\theequation}{\thesubsection.4}
\begin{equation}
\begin{aligned}
\mathcal{Q}(\boldsymbol{\Theta} \mid \boldsymbol{\Theta}') - H(\boldsymbol{\Theta} \mid \boldsymbol{\Theta}')
&=
\sum_{k=1}^K \sum_{\mathbf{I}_{CL}^k} 
\log 
\frac{\text{Pr}(\mathbf{t}^k, \mathbf{I}_{CL}^k \mid \boldsymbol{\Theta})}
{\text{Pr}(\mathbf{I}_{CL}^k \mid \mathbf{t}^k, \boldsymbol{\Theta})}
\text{Pr}(\mathbf{I}_{CL}^k \mid \mathbf{t}^k, \boldsymbol{\Theta}') \\
&=
\sum_{k=1}^K \sum_{\mathbf{I}_{CL}^k} 
\log \text{Pr}(\mathbf{t}^k \mid \boldsymbol{\Theta})
\text{Pr}(\mathbf{I}_{CL}^k \mid \mathbf{t}^k, \boldsymbol{\Theta}') \\
&=
\sum_{k=1}^K \log \text{Pr}(\mathbf{t}^k \mid \boldsymbol{\Theta}) \\
& =
\ell_{CL}(\boldsymbol{\Theta} \mid \mathbf{t}).
\end{aligned}
\end{equation}

\textbf{Lemma 2.}  
For any $(\boldsymbol{\Theta}, \boldsymbol{\Theta}') \in \Theta \times \Theta$,
\[
H(\boldsymbol{\Theta}' \mid \boldsymbol{\Theta}) 
\leq 
H(\boldsymbol{\Theta} \mid \boldsymbol{\Theta}).
\]

\noindent\emph{Proof.}
By Jensen’s inequality and the concavity of $\log$,
\renewcommand{\theequation}{\thesubsection.5}
\begin{equation}
\begin{aligned}
H(\boldsymbol{\Theta}' \mid \boldsymbol{\Theta}) - H(\boldsymbol{\Theta} \mid \boldsymbol{\Theta})
&=
\sum_{k=1}^K \sum_{\mathbf{I}_{CL}^k}
\log 
\frac{\text{Pr}(\mathbf{I}_{CL}^k \mid \mathbf{t}^k, \boldsymbol{\Theta}')}
{\text{Pr}(\mathbf{I}_{CL}^k \mid \mathbf{t}^k, \boldsymbol{\Theta})}
\text{Pr}(\mathbf{I}_{CL}^k \mid \mathbf{t}^k, \boldsymbol{\Theta}) \\
&\leq 
\log \sum_{k=1}^K 
\text{Pr}(\mathbf{I}_{CL}^k \mid \mathbf{t}^k, \boldsymbol{\Theta}')
= 0,
\end{aligned}
\end{equation}
which implies 
$H(\boldsymbol{\Theta}' \mid \boldsymbol{\Theta}) \leq H(\boldsymbol{\Theta} \mid \boldsymbol{\Theta})$.

\noindent\textbf{Proof of Theorem 1.}  
Given the CLEM sequence 
$\hat{\boldsymbol{\Theta}}^{(a)} 
= 
\arg\max_{\boldsymbol{\Theta}} 
\mathcal{Q}(\boldsymbol{\Theta} \mid \hat{\boldsymbol{\Theta}}^{(a-1)})$, 
for $a=1,2,\ldots$, 
we use Lemma~1 to write:
\[
\ell_{CL}(\hat{\boldsymbol{\Theta}}^{(a)} \mid \mathbf{t})
=
\mathcal{Q}(\hat{\boldsymbol{\Theta}}^{(a)} \mid \hat{\boldsymbol{\Theta}}^{(a-1)})
-
H(\hat{\boldsymbol{\Theta}}^{(a)} \mid \hat{\boldsymbol{\Theta}}^{(a-1)}).
\]
Since $\hat{\boldsymbol{\Theta}}^{(a)}$ maximizes $\mathcal{Q}(\boldsymbol{\Theta} \mid \hat{\boldsymbol{\Theta}}^{(a-1)})$, we have 
\[
\mathcal{Q}(\hat{\boldsymbol{\Theta}}^{(a)} \mid \hat{\boldsymbol{\Theta}}^{(a-1)})
\geq 
\mathcal{Q}(\hat{\boldsymbol{\Theta}}^{(a-1)} \mid \hat{\boldsymbol{\Theta}}^{(a-1)}).
\]
Meanwhile, by Lemma~2,
\[
H(\hat{\boldsymbol{\Theta}}^{(a)} \mid \hat{\boldsymbol{\Theta}}^{(a-1)}) 
\leq 
H(\hat{\boldsymbol{\Theta}}^{(a-1)} \mid \hat{\boldsymbol{\Theta}}^{(a-1)}).
\]
Combining these two inequalities yields
\renewcommand{\theequation}{\thesubsection.6}
\begin{equation}
\begin{aligned}
\ell_{CL}(\hat{\boldsymbol{\Theta}}^{(a)} \mid \mathbf{t})
&=
\mathcal{Q}(\hat{\boldsymbol{\Theta}}^{(a)} \mid \hat{\boldsymbol{\Theta}}^{(a-1)})
-
H(\hat{\boldsymbol{\Theta}}^{(a)} \mid \hat{\boldsymbol{\Theta}}^{(a-1)}) \\
&\geq
\mathcal{Q}(\hat{\boldsymbol{\Theta}}^{(a-1)} \mid \hat{\boldsymbol{\Theta}}^{(a-1)})
-
H(\hat{\boldsymbol{\Theta}}^{(a-1)} \mid \hat{\boldsymbol{\Theta}}^{(a-1)}) \\
&=
\ell_{CL}(\hat{\boldsymbol{\Theta}}^{(a-1)} \mid \mathbf{t}),
\end{aligned}
\end{equation}
which establishes the monotone ascent property in (\ref{eq: theorem1}).  
Equality holds if and only if 
$\mathcal{Q}(\hat{\boldsymbol{\Theta}}^{(a)} \mid \hat{\boldsymbol{\Theta}}^{(a-1)}) 
= 
\mathcal{Q}(\hat{\boldsymbol{\Theta}}^{(a-1)} \mid \hat{\boldsymbol{\Theta}}^{(a-1)})$.

\subsection{Proof of Statistical Consistency}
\label{Proof of Theorem 2}
\noindent\textbf{Statistical Consistency.} 
Suppose that the parameter space $\Theta$ is compact and the following conditions hold:

\begin{enumerate}[(1)]
\item $\mathbb{E}[U_{K,d}(\boldsymbol{\Theta})] = 0$ if and only if $\boldsymbol{\Theta} = \boldsymbol{\Theta}_d$;
\item There exists a nonnegative function $\tau(\cdot)$ such that
\renewcommand{\theequation}{\thesubsection.1}
\begin{equation}
    \left\|\frac{f^{(1)}(\mathbf{t}^k \mid \boldsymbol{\Theta})}{f(\mathbf{t}^k \mid \boldsymbol{\Theta})}\right\| < \tau(n^k)
\end{equation}
\renewcommand{\theequation}{\thesubsection.2}
\begin{equation}
\left\|\frac{f^{(2)}(\mathbf{t}^k \mid \boldsymbol{\Theta})}{f(\mathbf{t}^k \mid \boldsymbol{\Theta})}\right\| < \tau(n^k)
\end{equation}
where $\mathbb{E}[\tau(n^k)^2] < \infty$ and $n^k = |\mathbf{t}^k|$ denotes the number of events in the $k$th sub-window. 
Then $\hat{\boldsymbol{\Theta}}_{K,d}$ converges in probability to $\boldsymbol{\Theta}_d$ as $K \rightarrow \infty$. Furthermore, if $\mathbb{E}[U_{K,d}(\boldsymbol{\Theta})] \rightarrow 0$ as $d \rightarrow \infty$ only when $\boldsymbol{\Theta} = \boldsymbol{\Theta}_0$, then $\boldsymbol{\Theta}_d \rightarrow \boldsymbol{\Theta}_0$ as $d \rightarrow \infty$.
\end{enumerate}

\noindent To establish Theorem~2, we build upon the following lemmas.

\noindent\textbf{Lemma 3 (Theorem 3.1 in \citep{crowder1986consistency}).}
Let $\hat{\boldsymbol{\Theta}}_{K,d}$ be the solution to $U_{K,d}(\boldsymbol{\Theta}) = 0$. 
Define $B_\epsilon = \{\|\boldsymbol{\Theta} - \boldsymbol{\Theta}_d\| < \epsilon\}$ for some $\epsilon > 0$. 
Then $P[\hat{\boldsymbol{\Theta}}_{K,d} \in B_\epsilon] \rightarrow 1$ if the following conditions hold:
\begin{itemize}
    \item[(A1)] $\mathbb{E}[U_{K,d}(\boldsymbol{\Theta})] \rightarrow \boldsymbol{0}$ only at $\boldsymbol{\Theta} = \boldsymbol{\Theta}_d$;
    \item[(A2)] $\inf_{\Theta - B_\epsilon} \|\mathbb{E}[U_{K,d}(\boldsymbol{\Theta})]\| \ge C_\epsilon$ for some $C_\epsilon > 0$;
    \item[(A3)] $\sup_{\Theta - B_\epsilon} \|U_{K,d}(\boldsymbol{\Theta}) - \mathbb{E}[U_{K,d}(\boldsymbol{\Theta})]\| \rightarrow 0$ in probability.
\end{itemize}

\noindent\textbf{Lemma 4.}
If $\mathbb{E}[U_{K,d}(\boldsymbol{\Theta})]$ is continuous in $\boldsymbol{\Theta}$, then (A1) and (A2) in Lemma~3 are satisfied.

\textit{Proof.}  
Since $\mathbb{E}[U_{K,d}(\boldsymbol{\Theta})]$ is continuous and equals zero only at $\boldsymbol{\Theta} = \boldsymbol{\Theta}_d$, which implies (A1). 
Continuity also implies boundedness of $\mathbb{E}[U_{K,d}(\boldsymbol{\Theta})]$ on the compact space $\Theta$. 
Hence, there exists $C_\epsilon > 0$ such that $\inf_{\Theta - B_\epsilon}\|\mathbb{E}[U_{K,d}(\boldsymbol{\Theta})]\| \ge \inf_{\Theta}\|\mathbb{E}[U_{K,d}(\boldsymbol{\Theta})]\| \ge C_\epsilon$, verifying (A2). 

\vspace{1em}
Next, we prove the continuity of $\mathbb{E}[U_{K,d}(\boldsymbol{\Theta})]$ in $\Theta$ through Lemmas~5 and~6.

\noindent\textbf{Lemma 5.}
If there exists a nonnegative function $\tau(\cdot)$ satisfying 
\[
\Big\|\frac{f^{(1)}(\mathbf{t}^k|\boldsymbol{\Theta})}{f(\mathbf{t}^k|\boldsymbol{\Theta})}\Big\| < \tau(n^k), \quad 
\Big\|\frac{f^{(2)}(\mathbf{t}^k|\boldsymbol{\Theta})}{f(\mathbf{t}^k|\boldsymbol{\Theta})}\Big\| < \tau(n^k), \quad 
\mathbb{E}[\tau(n^k)^2] < \infty,
\]
then $\mathbb{E}[U'_{K,d}(\boldsymbol{\Theta})]$ is bounded.

\textit{Proof.}
Since 
\[
U'_{K,d}(\boldsymbol{\Theta}) = \sum_{k=1}^K \left[\frac{f^{(2)}(\mathbf{t}^k|\boldsymbol{\Theta})}{f(\mathbf{t}^k|\boldsymbol{\Theta})} 
- \Big(\frac{f^{(1)}(\mathbf{t}^k|\boldsymbol{\Theta})}{f(\mathbf{t}^k|\boldsymbol{\Theta})}\Big)^2 \right],
\]
we have
\renewcommand{\theequation}{\thesubsection.3}
\begin{equation}
  \begin{aligned}
\label{eq:lemma4}
 \mathbb{E}[U'_{K,d}(\boldsymbol{\Theta})] &  = \mathbb{E}[\sum_{k=1}^K(\frac{f^{(2)}(\mathbf{t}^k|\boldsymbol{\Theta})}{f(\mathbf{t}^k|\boldsymbol{\Theta})} - (\frac{f^{(1)}(\mathbf{t}^k|\boldsymbol{\Theta})}{f(\mathbf{t}^k|\boldsymbol{\Theta})})^2)]\\
  & \leq \sum_{k=1}^K \mathbb{E}[\|\frac{f^{(2)}(\mathbf{t}^k|\boldsymbol{\Theta})}{f(\mathbf{t}^k|\boldsymbol{\Theta})}\|] +  \sum_{k=1}^K \mathbb{E}[\|\frac{f^{(1)}(\mathbf{t}^k|\boldsymbol{\Theta})}{f(\mathbf{t}^k|\boldsymbol{\Theta})}\|^2]  \\ 
  & \leq \sum_{k=1}^K \mathbb{E}[\tau(n^k)] + \sum_{k=1}^K \mathbb{E}[\tau(n^k)^2] \\
  & < \infty,
  \end{aligned}
\end{equation}

Thus, $\mathbb{E}[U'_{K,d}(\boldsymbol{\Theta})]$ is bounded and Lemma 5 is proofed.

\vspace{1em}
\noindent\textbf{Lemma 6.}
If $U_{K,d}(\boldsymbol{\Theta})$ is continuous and differentiable on the compact set $\Theta$, and $\mathbb{E}[U'_{K,d}(\boldsymbol{\Theta})]$ is bounded, then $\mathbb{E}[U_{K,d}(\boldsymbol{\Theta})]$ is continuous on $\Theta$.

\textit{Proof.}
By the Mean Value Theorem, for any $\boldsymbol{\delta}$, there exists $\widetilde{\boldsymbol{\Theta}} \in (\boldsymbol{\Theta}, \boldsymbol{\Theta} + \boldsymbol{\delta})$ such that
\[
\mathbb{E}[U_{K,d}(\boldsymbol{\Theta} + \boldsymbol{\delta}) - U_{K,d}(\boldsymbol{\Theta})] 
= \mathbb{E}[U'_{K,d}(\widetilde{\boldsymbol{\Theta}})] \cdot \|\boldsymbol{\delta}\|.
\]
Given $\|\mathbb{E}[U'_{K,d}(\boldsymbol{\Theta})]\| < M$, for any $\epsilon^* > 0$, choose $\delta^* = \epsilon^*/M$. 
Then, whenever $\|\boldsymbol{\Theta} - \boldsymbol{\Theta}^*\| < \delta^*$,
\[
\|\mathbb{E}[U_{K,d}(\boldsymbol{\Theta})] - \mathbb{E}[U_{K,d}(\boldsymbol{\Theta}^*)]\| 
< \epsilon^*.
\]
Hence, $\mathbb{E}[U_{K,d}(\boldsymbol{\Theta})]$ is continuous. 

\vspace{1em}
Having established continuity, we have verified (A1) and (A2) in Lemma~3. 
Next, we verify (A3) using Lemmas~7–9.

\noindent\textbf{Lemma 7(\citep{guan2006composite}).} If for any $\epsilon>0, \eta>0$, there exist $\delta>0$ and $K'<\infty$ such that the following two conditions are satisfied for $K>K'$, we can  verify (A3) in Lemma 3.

(C1) $\sup_{\|\boldsymbol{\Theta}-\boldsymbol{\Theta}_d\|<\delta} \|\mathbb{E}[U_{K,d}(\boldsymbol{\Theta}_1)] - \mathbb{E}[U_{K,d}(\boldsymbol{\Theta}_2)]\|<\epsilon$,

(C2) $P[\sup_{\|\boldsymbol{\Theta}-\boldsymbol{\Theta}_d\|<\delta} \|U_{K,d}(\boldsymbol{\Theta}_1) - U_{K,d}(\boldsymbol{\Theta}_2)]\|]<\eta$.

We have proofed that $\mathbb{E}[U_{K,d}(\boldsymbol{\Theta})]$ is continuous on $\Theta$ in Lemma 6, which directly implies (C1) in Lemma 7. Thus to verify (A3) in \textbf{Lemma 3}, we only need to verify (C2) in \textbf{Lemma 7}. Now, we are going to verify (C2) by Lemma 8 and Lemma 9.

\noindent\textbf{Lemma 8.} If there exists a nonnegative function $\tau(\cdot)$ such that $\|\frac{f^{(1)}(\mathbf{t}^k|\boldsymbol{\Theta})}{f(\mathbf{t}^k|\boldsymbol{\Theta})}\|<\tau(n^k), \\ \|\frac{f^{(2)}(\mathbf{t}^k|\boldsymbol{\Theta})}{f(\mathbf{t}^k|\boldsymbol{\Theta})}\|<\tau(n^k)$, and $\mathbb{E}[\tau(n^k)^2]$ is bounded, then for any $\eta$, there exists $M_\eta <\infty$ such that $\text{Pr}[\sup_{\boldsymbol{\Theta}}\|U'_{K,d}(\boldsymbol{\Theta})\|>M_\eta]<\eta$.

\textit{Proof:} We have 
\renewcommand{\theequation}{\thesubsection.4}
\begin{equation}
  \begin{aligned}
\label{eq:lemma3}
 \|U'_{K, d}(\boldsymbol{\Theta})\| 
 & \leq \sum_{k=1}^K \|\frac{f^{(2)}(\mathbf{t}^k|\boldsymbol{\Theta})}{f(\mathbf{t}^k|\boldsymbol{\Theta})}\| + \sum_{k=1}^K\|\frac{f^{(1)}(\mathbf{t}^k|\boldsymbol{\Theta})}{f(\mathbf{t}^k|\boldsymbol{\Theta})}\|^2
 \\
  & \leq \sum_{k=1}^K[\tau(n^k) + \tau(n^k)^2].
  \end{aligned}
\end{equation}

Since $\tau(\cdot)$ is nonnegative and $\mathbb{E}[\tau(n^k)^2]$ is bounded, there exists $k_0<\infty$ such that $\sum_{k=1}^K \mathbb{E}[\tau(|n^k|) + \tau(|n^k|)^2]<k_0$. By the Markov inequality, we have 
\renewcommand{\theequation}{\thesubsection.5}
\begin{equation}
  \begin{aligned}
\label{eq:lemma3}
 \text{Pr}\{\sum_{k=1}^K [\tau(n^k) + \tau(n^k)^2] > \frac{k_0}{\eta}\}
 & \leq \frac{\sum_{k=1}^K \mathbb{E}[\tau(n^k) + \tau(n^k)^2]}{k_0/\eta}
 \\
  & \leq \eta.
  \end{aligned}
\end{equation}

Let $M_\eta = \frac{k_0}{\eta}$, we have 
\renewcommand{\theequation}{\thesubsection.6}
\begin{equation}
  \begin{aligned}
\label{eq:lemma3}
 \text{Pr}\{\sup_{\boldsymbol{\Theta}}\|U'_{K,d}(\boldsymbol{\Theta})\|>M_\eta\}
 & \leq \text{Pr}\{\sum_{k=1}^K [\tau(n^k) + \tau(n^k)^2]>M_\eta\}
 \\
  & \leq \eta.
  \end{aligned}
\end{equation}

\noindent\textbf{Lemma 9.} If for any $\eta>0$, there exists $M_{\eta}<\infty$, such that $\text{Pr}[\sup_{\boldsymbol{\Theta}}\|U'_{K,d}(\boldsymbol{\Theta})\|>M_{\eta}]<\eta$, then, $\text{Pr}[\sup_{\|\boldsymbol{\Theta}-\boldsymbol{\Theta}_s\|<\delta}\|U_{K,d}(\boldsymbol{\Theta_1})-U_{K,d}(\boldsymbol{\Theta_2})\|>\frac{\epsilon}{2}]<\eta$.

\textit{Proof:} By the continuity of  
$U_{K,d}(\boldsymbol{\Theta})$ and the Mean Value Theorem, we have
\renewcommand{\theequation}{\thesubsection.7}
\begin{equation}
  \begin{aligned}
\label{eq:lemma5}
 \|U_{K,d}(\boldsymbol{\Theta}_1) -U_{K,d}(\boldsymbol{\Theta}_2)\|_{\|\boldsymbol{\Theta}-\boldsymbol{\Theta}_s\|<\delta} &  = U'_{K,d}(\widetilde{\boldsymbol{\Theta}}) \cdot \|\boldsymbol{\Theta}_1-\boldsymbol{\Theta}_2\|\\
  & < \|U'_{K,d}(\widetilde{\boldsymbol{\Theta}})\|\cdot \delta,
  \end{aligned}
\end{equation}
which implies that $\sup_{\|\boldsymbol{\Theta}-\boldsymbol{\Theta}_s\|<\delta}\|U_{K,d}(\boldsymbol{\Theta}_1)-U_{K,d}(\boldsymbol{\Theta}_2)\|<\sup_{\boldsymbol{\Theta}}\|U'_{K,d}(\widetilde{\boldsymbol{\Theta}})\|\cdot \delta$. Thus, for any $\epsilon, \eta$, there exists $M_{\eta} = \frac{\epsilon}{2\delta}<\infty$ such that $\sup_{\|\boldsymbol{\Theta}-\boldsymbol{\Theta}_s\|<\delta}\|U_{K,d}(\boldsymbol{\Theta}_1) -U_{K,d}(\boldsymbol{\Theta}_2)\|>\frac{\epsilon}{2}$ implies $\sup_{\boldsymbol{\Theta}}\|U'_{K,d}(\widetilde{\boldsymbol{\Theta}})\|\cdot \delta>\frac{\epsilon}{2}$, which further implies $\text{Pr}(\sup_{\|\boldsymbol{\Theta}-\boldsymbol{\Theta}_s\|<\delta}\|U_{K,d}(\boldsymbol{\Theta}_1) -U_{K,d}(\boldsymbol{\Theta}_2)\|>\frac{\epsilon}{2}) < \text{Pr}(\sup_{\boldsymbol{\Theta}}\|U'_{K,d}(\widetilde{\boldsymbol{\Theta}})\|>\frac{\epsilon}{2\delta}=M_\eta)<\eta$. Thus, we proofed Lemma 9., i.e., (C2) in Lemma 7.

Thus far, we verified (A1)-(A3) in Lemma 3, which implies $\hat{\boldsymbol{\Theta}}_{K,d}$ converges in probability to $\boldsymbol{\Theta}_d$ as $K\rightarrow \infty$. Next, we are going to proof that if $\mathbb{E}[U_{K,d}(\boldsymbol{\Theta})] \rightarrow 0$ as $d \rightarrow \infty$ only at $\boldsymbol{\Theta} = \boldsymbol{\Theta}_0$, we have $\boldsymbol{\Theta}_d \rightarrow \boldsymbol{\Theta}_0$ as $d \rightarrow \infty$ in Lemma 10.

\noindent \textbf{Lemma 10.} If $\hat{\boldsymbol{\Theta}}_{K,d}$ converges in probability to $\boldsymbol{\Theta}_d$ as $K \rightarrow \infty$, then for any $\epsilon>0$, there exists $d'\in\mathbb{R}$ such that for any $d>d'$, we have 
\renewcommand{\theequation}{\thesubsection.8}
\begin{equation}
    \|\boldsymbol{\Theta}_d - \boldsymbol{\Theta}_0\|_{\infty}<\epsilon.
\end{equation}

\textit{Proof:} If the above statement is not true, then there exists $\epsilon^* \in \mathbb{R}$ such that for any $d'>0$, there exists $\widetilde{d}>d'$, such that $||\boldsymbol{\Theta}_d - \boldsymbol{\Theta}_0||_\infty>\epsilon^*$. Therefore, there exists a sequence $\{\boldsymbol{\Theta}^*_{\widetilde{d}_n}\}_{n=1}^\infty$ such that $\widetilde{d}_n \rightarrow \infty$ and $||\boldsymbol{\Theta}^*_{\widetilde{d}_n} - \boldsymbol{\Theta}^*_0||_\infty > \epsilon^*$. Since $\Theta$ is compact, there exists a subsequence $\{\boldsymbol{\Theta}^*_{\widetilde{d}_{n_k}}\}_{k=1}^\infty$ such that $\boldsymbol{\Theta}^*_{\widetilde{d}_{n_k}}\rightarrow\boldsymbol{\Theta}^*_0 \neq \boldsymbol{\Theta}_0$ as $k \rightarrow \infty$ and $||\boldsymbol{\Theta}^*_0 - \boldsymbol{\Theta}_0||_\infty > \epsilon^*$.

Define $\Phi_d(\boldsymbol{\Theta}) = \mathbb{E}[U_{K, d}(\boldsymbol{\Theta})]$. Since $U_{K, d}(\boldsymbol{\Theta})$ is continuous and differentiable, according to the Mean Value Theorem, we have 
\renewcommand{\theequation}{\thesubsection.9}
\begin{equation}
  \begin{aligned}
\label{eq:lemma3}
 \|\Phi_{\widetilde{d}_{n_k}}(\boldsymbol{\Theta}^*_{\widetilde{d}_{n_k}}) - \Phi_{\widetilde{d}_{n_k}}(\boldsymbol{\Theta}^*_0)\| &  = \|\Phi'_{\widetilde{d}_{n_k}}(\widetilde{\boldsymbol{\Theta}}_0)\| \cdot \|\boldsymbol{\Theta}^*_{\widetilde{d}_{n_k}} - \boldsymbol{\Theta}^*_0\| \\
  & \leq M \cdot \|\boldsymbol{\Theta}^*_{\widetilde{d}_{n_k}} - \boldsymbol{\Theta}^*_0\|.
  \end{aligned}
\end{equation}

We know $\Phi_{\widetilde{d}_{n_k}}(\boldsymbol{\Theta}^*_{\widetilde{d}_{n_k}}) = \mathbb{E}[U_{K, {\widetilde{d}_{n_k}}}(\boldsymbol{\Theta}^*_{\widetilde{d}_{n_k}})] \rightarrow 0$ as $\widetilde{d}_{n_k}\rightarrow\infty$ and $\Phi'_{\widetilde{d}_{n_k}}(\boldsymbol{\Theta})$ is bounded and continuous. Then, $\lim_{\widetilde{d}_{n_k}\rightarrow\infty}\|\Phi_{\widetilde{d}_{n_k}}(\boldsymbol{\Theta}^*_{\widetilde{d}_{n_k}}) - \Phi_{\widetilde{d}_{n_k}}(\boldsymbol{\Theta}^*_0)\| \leq M \cdot \lim_{\widetilde{d}_{n_k}\rightarrow\infty}\|\boldsymbol{\Theta}^*_{\widetilde{d}_{n_k}} - \boldsymbol{\Theta}^*_0\|$ and $\lim_{\widetilde{d}_{n_k}\rightarrow\infty}\| \Phi_{\widetilde{d}_{n_k}}(\boldsymbol{\Theta}^*_0)\| \leq M \cdot \|\boldsymbol{\Theta}^*_0 - \boldsymbol{\Theta}^*_0\| = 0$, which implies $\lim_{\widetilde{d}_{n_k}\rightarrow\infty} \Phi_{\widetilde{d}_{n_k}}(\boldsymbol{\Theta}^*_0) = 0$. This is contradicted with the assumption that $\lim_{d\rightarrow\infty} \Phi_d(\boldsymbol{\Theta}) = 0$ only at $\boldsymbol{\Theta} = \boldsymbol{\Theta}_0$. Thus, we have proved Lemma 10. 

\section{Appendix: Additional Results in Section 3.1.4}
\label{app:III}

Tables~\ref{tab:Friedman test T1000}, \ref{tab:Friedman test T2500}, and \ref{tab:Friedman test T5000} report the stepwise Friedman test results for $T = 1000$, $2500$, and $5000$. Using the same selection criterion, we obtain the following optimal choices: $K^* = 20$ for $T = 1000$, $K^* = 50$ for $T = 2500$, and $K^* = 100$ for $T = 5000$. 
\begin{table*}[h]
\small
\centering
\begin{threeparttable}
\caption{Stepwise Friedman Test Results for Selecting the Optimal $K$ ($T=1000$)}
\label{tab:Friedman test T1000}
\begin{tabular}{lll}
\hline
\hline
\textbf{Hypothesis} & \textbf{$p$-value} & \textbf{Conclusion}  \\
\hline
$H_0^{(1)}: M_{EM} = M_{CLEM(2)} = M_{CLEM(5)}$ & \multirow{2}{*}{0.756
} & \multirow{2}{*}{Not reject $H_0^1$}   \\
$H_1^{(1)}: \text{Not all medians are equal}$ &  &   \\
\hline
$H_0^{(2)}: M_{EM} = M_{CLEM(2)} = M_{CLEM(5)} =  M_{CLEM(10)}$ & \multirow{2}{*}{0.147

} & \multirow{2}{*}{Not reject $H_0^2$}   \\
$H_1^{(2)}: \text{Not all medians are equal}$ &  &   \\
\hline
$H_0^{(3)}: M_{EM} = M_{CLEM(2)} = M_{CLEM(5)} =  M_{CLEM(10)} =  M_{CLEM(20)} $ & \multirow{2}{*}{0.137
} & \multirow{2}{*}{Not reject $H_0^3$}   \\
$H_1^{(3)}: \text{Not all medians are equal}$ &  &   \\
\hline
$H_0^{(4)}: M_{EM} = M_{CLEM(2)} = M_{CLEM(5)} = \cdots =  M_{CLEM(50)} $ & \multirow{2}{*}{$<0.001$
} & \multirow{2}{*}{\textbf{Reject $H_0^4$}}  \\
$H_1^{(4)}: \text{Not all medians are equal}$ &  &   \\
\hline
\hline
\end{tabular}

\begin{tablenotes}
\item[a] Rejection decision is made if $p$-value is smaller that 0.05.
\end{tablenotes}
\end{threeparttable}
\end{table*}

\begin{table*}[!t]
\small
\centering
\begin{threeparttable}
\caption{Stepwise Friedman Test Results for Selecting the Optimal $K$ ($T=2500$)}
\label{tab:Friedman test T2500}
\begin{tabular}{lll}
\hline
\hline
\textbf{Hypothesis} & \textbf{$p$-value} & \textbf{Conclusion}  \\
\hline
$H_0^{(1)}: M_{EM} = M_{CLEM(2)} = M_{CLEM(5)}$ & \multirow{2}{*}{0.533
} & \multirow{2}{*}{Not reject $H_0^1$}   \\
$H_1^{(1)}: \text{Not all medians are equal}$ &  &   \\
\hline
$H_0^{(2)}: M_{EM} = M_{CLEM(2)} = M_{CLEM(5)} =  M_{CLEM(10)}$ & \multirow{2}{*}{0.705

} & \multirow{2}{*}{Not reject $H_0^2$}   \\
$H_1^{(2)}: \text{Not all medians are equal}$ &  &   \\
\hline
$H_0^{(3)}: M_{EM} = M_{CLEM(2)} = M_{CLEM(5)} =  M_{CLEM(10)} =  M_{CLEM(20)} $ & \multirow{2}{*}{0.846
} & \multirow{2}{*}{Not reject $H_0^3$}   \\
$H_1^{(3)}: \text{Not all medians are equal}$ &  &   \\
\hline
$H_0^{(4)}: M_{EM} = M_{CLEM(2)} = M_{CLEM(5)} = \cdots =  M_{CLEM(50)} $ & \multirow{2}{*}{0.106
} & \multirow{2}{*}{Not reject $H_0^4$}   \\
$H_1^{(4)}: \text{Not all medians are equal}$ &  &   \\
\hline
$H_0^{(5)}: M_{EM} = M_{CLEM(2)} = M_{CLEM(5)} = \cdots =  M_{CLEM(100)} $ & \multirow{2}{*}{$<0.001$
} & \multirow{2}{*}{\textbf{Reject $H_0^5$}}   \\
$H_1^{(5)}: \text{Not all medians are equal}$ &  &   \\
\hline
\hline
\end{tabular}
\begin{tablenotes}
\item[a] Rejection decision is made if $p$-value is smaller that 0.05.
\end{tablenotes}
\end{threeparttable}
\end{table*}

\begin{table*}[!t]
\small
\centering
\begin{threeparttable}
\caption{Stepwise Friedman Test Results for Selecting the Optimal $K$ ($T=5000$)}
\label{tab:Friedman test T5000}
\begin{tabular}{lll}
\hline
\hline
\textbf{Hypothesis} & \textbf{$p$-value} & \textbf{Conclusion}  \\
\hline
$H_0^{(1)}: M_{EM} = M_{CLEM(2)} = M_{CLEM(5)}$ & \multirow{2}{*}{0.210
} & \multirow{2}{*}{Not reject $H_0^1$}   \\
$H_1^{(1)}: \text{Not all medians are equal}$ &  &   \\
\hline
$H_0^{(2)}: M_{EM} = M_{CLEM(2)} = M_{CLEM(5)} =  M_{CLEM(10)}$ & \multirow{2}{*}{0.080

} & \multirow{2}{*}{Not reject $H_0^2$}   \\
$H_1^{(2)}: \text{Not all medians are equal}$ &  &   \\
\hline
$H_0^{(3)}: M_{EM} = M_{CLEM(2)} = M_{CLEM(5)} =  M_{CLEM(10)} =  M_{CLEM(20)} $ & \multirow{2}{*}{0.126
} & \multirow{2}{*}{Not reject $H_0^3$}   \\
$H_1^{(3)}: \text{Not all medians are equal}$ &  &   \\
\hline
$H_0^{(4)}: M_{EM} = M_{CLEM(2)} = M_{CLEM(5)} = \cdots =  M_{CLEM(50)} $ & \multirow{2}{*}{0.108
} & \multirow{2}{*}{Not reject $H_0^4$}   \\
$H_1^{(4)}: \text{Not all medians are equal}$ &  &   \\
\hline
$H_0^{(5)}: M_{EM} = M_{CLEM(2)} = M_{CLEM(5)} = \cdots =  M_{CLEM(100)} $ & \multirow{2}{*}{0.107
} & \multirow{2}{*}{Not reject $H_0^5$}   \\
$H_1^{(5)}: \text{Not all medians are equal}$ &  &   \\
\hline
$H_0^{(6)}: M_{EM} = M_{CLEM(2)} = M_{CLEM(5)} = \cdots =  M_{CLEM(250)} $ & \multirow{2}{*}{$<0.01$
} & \multirow{2}{*}{\textbf{Reject $H_0^6$}}   \\
$H_1^{(6)}: \text{Not all medians are equal}$ &  &   \\
\hline
\hline
\end{tabular}
\begin{tablenotes}
\item[a] Rejection decision is made if $p$-value is smaller that 0.05.
\end{tablenotes}
\end{threeparttable}
\end{table*}

\FloatBarrier
\clearpage
\bibliography{Literature.bib}

@article{kalra2016driving,
  title={Driving to safety: How many miles of driving would it take to demonstrate autonomous vehicle reliability?},
  author={Kalra, Nidhi and Paddock, Susan M},
  journal={Transportation Research Part A: Policy and Practice},
  volume={94},
  pages={182--193},
  year={2016},
  publisher={Elsevier}
}

@article{geiger2013vision,
  title={Vision meets robotics: The kitti dataset},
  author={Geiger, Andreas and Lenz, Philip and Stiller, Christoph and Urtasun, Raquel},
  journal={The International Journal of Robotics Research},
  volume={32},
  number={11},
  pages={1231--1237},
  year={2013},
  publisher={Sage Publications Sage UK: London, England}
}

@article{hossain1993estimating,
  title={Estimating the parameters of a non-homogeneous Poisson-process model for software reliability},
  author={Hossain and Syed A and Dahiya, Ram C},
  journal={IEEE Transactions on Reliability},
  volume={42},
  number={4},
  pages={604--612},
  year={1993},
  publisher={IEEE}
}

@inproceedings{musa1984logarithmic,
  title={A logarithmic Poisson execution time model for software reliability measurement},
  author={Musa, John D and Okumoto, Kazuhira},
  booktitle={Proceedings of the 7th international conference on Software engineering},
  pages={230--238},
  year={1984},
  organization={Citeseer}
}

@article{pyke1961markov,
  title={Markov renewal processes: definitions and preliminary properties},
  author={Pyke, Ronald},
  journal={The Annals of Mathematical Statistics},
  pages={1231--1242},
  year={1961},
  publisher={JSTOR}
}

@article{lewis1979simulation,
  title={Simulation of nonhomogeneous Poisson processes by thinning},
  author={Lewis, PA W and Shedler, Gerald S},
  journal={Naval research logistics quarterly},
  volume={26},
  number={3},
  pages={403--413},
  year={1979},
  publisher={Wiley Online Library}
}

@article{hassaballah2020vehicle,
  title={Vehicle detection and tracking in adverse weather using a deep learning framework},
  author={Hassaballah, Mahmoud and Kenk, Mourad A and Muhammad, Khan and Minaee, Shervin},
  journal={IEEE transactions on intelligent transportation systems},
  volume={22},
  number={7},
  pages={4230--4242},
  year={2020},
  publisher={IEEE}
}

@article{kilic2021lidar,
  title={Lidar light scattering augmentation Physics-based simulation of adverse weather conditions for 3d object detection},
  author={Kilic, Velat and Hegde, Deepti and Sindagi, Vishwanath and Cooper, A Brinton and Foster, Mark A and Patel, Vishal M},
  journal={arXiv preprint arXiv:2107.07004},
  year={2021}
}

@inproceedings{pan2022quantifying,
  title={Quantifying Error Propagation in Multi-Stage Perception System of Autonomous Vehicles via Physics-Based Simulation},
  author={Pan, Fenglian and Zhang, Yinwei and Head, Larry and Liu, Jian and Elli, Maria and Alvarez, Ignacio},
  booktitle={2022 Winter Simulation Conference (WSC)},
  pages={2511--2522},
  year={2022},
  organization={IEEE}
}

@INPROCEEDINGS{8987509,
  author={Zhao, Xingyu and Robu, Valentin and Flynn, David and Salako, Kizito and Strigini, Lorenzo},
  booktitle={2019 IEEE 30th International Symposium on Software Reliability Engineering (ISSRE)}, 
  title={Assessing the Safety and Reliability of Autonomous Vehicles from Road Testing}, 
  year={2019},
  volume={},
  number={},
  pages={13-23},
  doi={10.1109/ISSRE.2019.00012}}

@article{van2012estimation,
  title={On Estimation of the Intensity Function of a Point Process},
  author={Van Lieshout, Marie-Colette NM},
  journal={Methodology and Computing in Applied Probability},
  volume={14},
  number={3},
  pages={567--578},
  year={2012}
}

@article{pham2003nhpp,
  title={NHPP software reliability and cost models with testing coverage},
  author={Pham, Xuemei and Zhang},
  journal={European Journal of Operational Research},
  volume={145},
  number={2},
  pages={443--454},
  year={2003},
  publisher={Elsevier}
}

@article{veen2008estimation,
  title={Estimation of space--time branching process models in seismology using an EM--type algorithm},
  author={Veen, Alejandro and Schoenberg, Frederic P},
  journal={Journal of the American Statistical Association},
  volume={103},
  number={482},
  pages={614--624},
  year={2008},
  publisher={Taylor \& Francis}
}

@article{zou2022modeling,
  title={Modeling public acceptance of private autonomous vehicles: Value of time and motion sickness viewpoints},
  author={Zou, Xin and Logan, David B and Vu, Hai L},
  journal={Transportation Research Part C: Emerging Technologies},
  volume={137},
  pages={103548},
  year={2022},
  publisher={Elsevier}
}

@article{li2023modified,
  title={Modified DDPG car-following model with a real-world human driving experience with CARLA simulator},
  author={Li, Dianzhao and Okhrin, Ostap},
  journal={Transportation research part C: emerging technologies},
  volume={147},
  pages={103987},
  year={2023},
  publisher={Elsevier}
}

@article{crash,
	chapter = {Technology},
	title = {Self-{Driving} {Uber} {Car} {Kills} {Pedestrian} in {Arizona}, {Where} {Robots} {Roam}},
	issn = {0362-4331},
	url = {https://spectrum.ieee.org/google-has-spent-over-11-billion-on-selfdriving-tech},
	language = {en-US},
	urldate = {2022-04-28},
	journal = {The New York Times},
	author = {NHTSA},
	month = mar,
	year = {2017}
	}

@article{crowder1986consistency,
  title={On consistency and inconsistency of estimating equations},
  author={Crowder, Martin},
  journal={Econometric Theory},
  volume={2},
  number={3},
  pages={305--330},
  year={1986},
  publisher={Cambridge University Press}
}

@article{guan2006composite,
  title={A composite likelihood approach in fitting spatial point process models},
  author={Guan, Yongtao},
  journal={Journal of the American Statistical Association},
  volume={101},
  number={476},
  pages={1502--1512},
  year={2006},
  publisher={Taylor \& Francis}
}

@article{peres2020industrial,
  title={Industrial artificial intelligence in industry 4.0-systematic review, challenges and outlook},
  author={Peres, Ricardo Silva and Jia, Xiaodong and Lee, Jay and Sun, Keyi and Colombo, Armando Walter and Barata, Jose},
  journal={IEEE access},
  volume={8},
  pages={220121--220139},
  year={2020},
  publisher={IEEE}
}

@article{mnyakin2023applications,
  title={Applications of ai, iot, and cloud computing in smart transportation: A review},
  author={Mnyakin, Maxim},
  journal={Artificial Intelligence in Society},
  volume={3},
  number={1},
  pages={9--27},
  year={2023}
}

@inproceedings{chan2019survey,
  title={Survey of AI in cybersecurity for information technology management},
  author={Chan, Leong and Morgan, Ian and Simon, Hayden and Alshabanat, Fares and Ober, Devin and Gentry, James and Min, David and Cao, Renzhi},
  booktitle={2019 IEEE technology \& engineering management conference (TEMSCON)},
  pages={1--8},
  year={2019},
  organization={IEEE}
}

@article{balagurunathan2021requirements,
  title={Requirements and reliability of AI in the medical context},
  author={Balagurunathan, Yoganand and Mitchell, Ross and El Naqa, Issam},
  journal={Physica Medica},
  volume={83},
  pages={72--78},
  year={2021},
  publisher={Elsevier}
}

@misc{AIIncidentDB,
 author= {{AI~Incident}},
 year={2024},
 howpublished = {[{Online}]. {Artificial Intelligence Incident Database:}
 \url{https://incidentdatabase.ai}, accessed: December 26, 2024.
 }
}

@article{pan2024reliability,
  title={Reliability modeling for perception systems in autonomous vehicles: A recursive event-triggering point process approach},
  author={Pan, Fenglian and Zhang, Yinwei and Liu, Jian and Head, Larry and Elli, Maria and Alvarez, Ignacio},
  journal={Transportation Research Part C: Emerging Technologies},
  volume={169},
  pages={104868},
  year={2024},
  publisher={Elsevier}
}

@article{Faddietal2024,
author = {Faddi, Zakaria and {da~Mata}, Karen and Silva, Priscila and Nagaraju, Vidhyashree and Ghosh, Susmita and Kul, Gokhan and Fiondella, Lance},
title = {Quantitative assessment of machine learning reliability and resilience},
journal = {Risk Analysis, in press},
year ={2024}
}

@ARTICLE{Lianetal2021Robustness,
  AUTHOR =       {Jiayi Lian and Laura Freeman and Yili Hong and Xinwei Deng},
  TITLE =        {Robustness with Respect to Class Imbalance in Artificial Intelligence Classification Algorithms},
  JOURNAL =      {Journal of Quality Technology},
  volume ={53},
  pages = {505--525},
  YEAR =         {2021},
}

@article{wang2020deep,
  title={A deep learning based data fusion method for degradation modeling and prognostics},
  author={Wang, Feng and Du, Juan and Zhao, Yang and Tang, Tao and Shi, Jianjun},
  journal={IEEE Transactions on Reliability},
  volume={70},
  number={2},
  pages={775--789},
  year={2020},
  publisher={IEEE}
}

@misc{CAdriving,
author={{California Department of Motor Vehicles}},
title={Autonomous Vehicle Tester Program},
year ={2024},
howpublished={[{Online}]. {Available at:}
\url{https://www.dmv.ca.gov/portal/vehicle-industry-services/autonomous-vehicles/},
accessed: September 01, 2024.}
}

@inproceedings{gorjian2010review,
  title={A review on reliability models with covariates},
  author={Gorjian, Nima and Ma, Lin and Mittinty, Murthy and Yarlagadda, Prasad and Sun, Yong},
  booktitle={Engineering Asset Lifecycle Management: Proceedings of the 4th World Congress on Engineering Asset Management (WCEAM 2009), 28-30 September 2009},
  pages={385--397},
  year={2010},
  organization={Springer}
}

@article{li2017bayesian,
  title={Bayesian nonparametric modeling of heterogeneous time-to-event data with an unknown number of sub-populations},
  author={Li, Mingyang and Han, Jiali and Liu, Jian},
  journal={IISE Transactions},
  volume={49},
  number={5},
  pages={481--492},
  year={2017},
  publisher={Taylor \& Francis}
}

@article{li2016bayesian,
  title={Bayesian hazard modeling based on lifetime data with latent heterogeneity},
  author={Li, Mingyang and Liu, Jian},
  journal={Reliability Engineering \& System Safety},
  volume={145},
  pages={183--189},
  year={2016},
  publisher={Elsevier}
}

@article{zhang2018degradation,
  title={Degradation data analysis and remaining useful life estimation: A review on Wiener-process-based methods},
  author={Zhang, Zhengxin and Si, Xiaosheng and Hu, Changhua and Lei, Yaguo},
  journal={European Journal of Operational Research},
  volume={271},
  number={3},
  pages={775--796},
  year={2018},
  publisher={Elsevier}
}

@article{pan2011reliability,
  title={Reliability modeling of degradation of products with multiple performance characteristics based on gamma processes},
  author={Pan, Zhengqiang and Balakrishnan, Narayanaswamy},
  journal={Reliability Engineering \& System Safety},
  volume={96},
  number={8},
  pages={949--957},
  year={2011},
  publisher={Elsevier}
}

@article{yuan2019reliability,
  title={A reliability analysis method of accelerated performance degradation based on bayesian strategy},
  author={Yuan, Rong and Tang, Mao and Wang, Hui and Li, Haiqing},
  journal={IEEE Access},
  volume={7},
  pages={169047--169054},
  year={2019},
  publisher={IEEE}
}

@book{kapur2014reliability,
  title={Reliability engineering},
  author={Kapur, Kailash C and Pecht, Michael G},
  year={2014},
  publisher={John Wiley \& Sons}
}

@article{dempster1977maximum,
  title={Maximum likelihood estimation from incomplete data via the EM algorithm},
  author={Dempster, Arthur},
  journal={Journal of the Royal Statistical Society},
  volume={39},
  pages={1--38},
  year={1977}
}

@inproceedings{okamura2003iterative,
  title={An iterative scheme for maximum likelihood estimation in software reliability modeling},
  author={Okamura, Hiroyuki and Watanabe, Yasuhiro and Dohi, Tadashi},
  booktitle={14th International Symposium on Software Reliability Engineering, 2003. ISSRE 2003.},
  pages={246--256},
  year={2003},
  organization={IEEE}
}

@article{okamura2013application,   title={Application of EM algorithm to NHPP-based software reliability assessment with ungrouped failure time data},   author={Okamura, Hiroyuki and Dohi, Tadashi},   journal={Stochastic Reliability and Maintenance Modeling: Essays in Honor of Professor Shunji Osaki on his 70th Birthday},   pages={285--313},   year={2013},   publisher={Springer} }

@article{zeephongsekul2016maximum,
  title={Maximum-likelihood estimation of parameters of NHPP software reliability models using expectation conditional maximization algorithm},
  author={Zeephongsekul, Panlop and Jayasinghe, Chathuri L and Fiondella, Lance and Nagaraju, Vidhyashree},
  journal={IEEE Transactions on Reliability},
  volume={65},
  number={3},
  pages={1571--1583},
  year={2016},
  publisher={IEEE}
}

@inproceedings{quigley2009ros,
  title={ROS: an open-source Robot Operating System},
  author={Quigley, Morgan and Conley, Ken and Gerkey, Brian and Faust, Josh and Foote, Tully and Leibs, Jeremy and Wheeler, Rob and Ng, Andrew Y and others},
  booktitle={ICRA workshop on open source software},
  volume={3},
  number={3.2},
  pages={5},
  year={2009},
  organization={Kobe, Japan}
}

@book{cox1980point,
  title={Point processes},
  author={Cox, David Roxbee and Isham, Valerie},
  volume={12},
  year={1980},
  publisher={CRC Press}
}

@article{xia2025penalized,
  title={Penalized spatial-temporal sensor fusion for detecting and localizing bursts in water distribution systems},
  author={Xia, Shenghao and Zhang, Yinwei and Lansey, Kevin and Liu, Jian},
  journal={Information Fusion},
  volume={117},
  pages={102912},
  year={2025},
  publisher={Elsevier}
}

@article{min2022reliability,
  title={Reliability analysis of artificial intelligence systems using recurrent events data from autonomous vehicles},
  author={Min, Jie and Hong, Yili and King, Caleb B and Meeker, William Q},
  journal={Journal of the Royal Statistical Society Series C: Applied Statistics},
  volume={71},
  number={4},
  pages={987--1013},
  year={2022},
  publisher={Oxford University Press}
}

@article{zheng2025dr,
  title={DR-AIR: A data repository bridging the research gap in AI reliability},
  author={Zheng, Simin and Clark, Jared M and Salboukh, Fatemeh and Silva, Priscila and da Mata, Karen and Pan, Fenglian and Min, Jie and Lian, Jiayi and King, Caleb B and Fiondella, Lance and others},
  journal={Quality Engineering},
  pages={1--22},
  year={2025},
  publisher={Taylor \& Francis}
}

@article{pan2025modeling,
  title={Modeling opioid overdose events recurrence with a covariate-adjusted triggering point process},
  author={Pan, Fenglian and Zhou, You and Vivas-Valencia, Carolina and Kong, Nan and Ott, Carol and Jalali, Mohammad S and Liu, Jian},
  journal={PLOS Computational Biology},
  volume={21},
  number={5},
  pages={e1012889},
  year={2025},
  publisher={Public Library of Science San Francisco, CA USA}
}

@article{kafka2012automotive,
  title={The automotive standard ISO 26262, the innovative driver for enhanced safety assessment \& technology for motor cars},
  author={Kafka, Peter},
  journal={Procedia Engineering},
  volume={45},
  pages={2--10},
  year={2012},
  publisher={Elsevier}
}

@article{yang2024measurement,
  title={Measurement error--tolerant Poisson regression for Valley Fever incidence prediction},
  author={Yang, Haomiao and Pan, Fenglian and Tong, Daoqin and Brown, Heidi E and Liu, Jian},
  journal={IISE transactions on healthcare systems engineering},
  volume={14},
  number={4},
  pages={305--317},
  year={2024},
  publisher={Taylor \& Francis}
}

@article{sheldon1996use,
  title={The use and interpretation of the Friedman test in the analysis of ordinal-scale data in repeated measures designs},
  author={Sheldon, Michael R and Fillyaw, Michael J and Thompson, W Douglas},
  journal={Physiotherapy Research International},
  volume={1},
  number={4},
  pages={221--228},
  year={1996},
  publisher={Wiley Online Library}
}

@article{liang2003maximum,
  title={Maximum pseudo likelihood estimation in network tomography},
  author={Liang, Gang and Yu, Bin},
  journal={IEEE Transactions on Signal Processing},
  volume={51},
  number={8},
  pages={2043--2053},
  year={2003},
  publisher={IEEE}
}

@article{varin2005pairwise,
  title={Pairwise likelihood inference in spatial generalized linear mixed models},
  author={Varin, Cristiano and H{\o}st, Gudmund and Skare, {\O}ivind},
  journal={Computational statistics \& data analysis},
  volume={49},
  number={4},
  pages={1173--1191},
  year={2005},
  publisher={Elsevier}
}

@article{gao2011composite,
  title={Composite likelihood EM algorithm with applications to multivariate hidden Markov model},
  author={Gao, Xin and Song, Peter X-K},
  journal={Statistica Sinica},
  pages={165--185},
  year={2011},
  publisher={JSTOR}
}

@article{ohishi2009gompertz,
  title={Gompertz software reliability model: Estimation algorithm and empirical validation},
  author={Ohishi, Koji and Okamura, Hiroyuki and Dohi, Tadashi},
  journal={Journal of Systems and software},
  volume={82},
  number={3},
  pages={535--543},
  year={2009},
  publisher={Elsevier}
}

@article{sahinoglu1992compound,
  title={Compound-Poisson software reliability model},
  author={Sahinoglu, Mehmet},
  journal={IEEE Transactions on Software Engineering},
  volume={18},
  number={7},
  pages={624},
  year={1992},
  publisher={IEEE Computer Society}
}
\bibliographystyle{apalike}
\end{document}